\documentclass[conference]{IEEEtran}
\usepackage{fancyhdr}

\AtBeginDocument{%
  \providecommand\BibTeX{{%
    \normalfont B\kern-0.5em{\scshape i\kern-0.25em b}\kern-0.8em\TeX}}}

\usepackage[utf8]{inputenc}
\usepackage{amsfonts}

\usepackage{amsmath}
\DeclareMathOperator*{\argmax}{arg\,max}

\usepackage{algorithm}
\usepackage[noend]{algpseudocode}
\usepackage{algorithmicx}
\renewcommand{\algorithmicrequire}{\textbf{Input:}}
\renewcommand{\algorithmicensure}{\textbf{Output:}}
\renewcommand{\algorithmicindent}{12pt}
\usepackage{clrs+,listings, url}
\usepackage{hyperref,xcolor}

\usepackage{graphicx, color}

\begin{document}

\title{Reducing Communication in Graph Neural Network Training}

\author{\IEEEauthorblockN{
     Alok Tripathy,
     Katherine Yelick,
     Ayd{\i}n Bulu\c{c}}    \vspace{0.5em}
   \IEEEauthorblockA{\IEEEauthorrefmark{1}
      Electrical Engineering and Computer Sciences,
     University of California, Berkeley} 
   \IEEEauthorblockA{\IEEEauthorrefmark{2}
     Computational Research Division,
     Lawrence Berkeley National Laboratory}
}  
\maketitle
\thispagestyle{fancy}
\lhead{}
\rhead{}
\chead{}
\lfoot{\footnotesize{
SC20, November 9-19, 2020, Is Everywhere We Are
\newline 978-1-7281-9998-6/20/\$31.00 \copyright 2020 IEEE}}
\rfoot{}
\cfoot{}
\renewcommand{\headrulewidth}{0pt}
\renewcommand{\footrulewidth}{0pt}

\newcommand{\mA}{\mathbf{A}} 
\newcommand{\mZ}{\mathbf{Z}}
\newcommand{\mG}{\mathbf{G}}
\newcommand{\transpose}     {^{\mbox{\scriptsize \sf T}}}
\newcommand{\mB}{\mathbf{B}}
\newcommand{\mW}{\mathbf{W}}
\newcommand{\mY}{\mathbf{Y}}
\newcommand{\mH}{\mathbf{H}}
\newcommand{\mT}{\mathbf{T}}
\newcommand{\dnnz}{\mathit{nnz}}

\newcommand{\revision}[1]{\textcolor{black}{#1}}
\newcommand{\revisiontwo}[1]{\textcolor{black}{#1}}
\newcommand{\hide}[1]{}
\newcommand{\Aydin}[1]{{\color{red}(Aydin) #1}}
\newcommand{\Alok}[1]{{\color{magenta}(Alok) #1}}
\newcommand{\Kathy}[1]{{\color{cyan}(Kathy) #1}}
\newcommand{\todo}[1]{{\color{green}(Todo:) #1}}

\begin{abstract}
Graph Neural Networks (GNNs) are powerful and flexible neural networks that use the naturally sparse connectivity information of the data. GNNs represent this connectivity as sparse matrices, which have lower arithmetic intensity and thus higher communication costs compared to dense matrices, making GNNs harder to scale to high concurrencies than convolutional or fully-connected neural networks.

We introduce a family of parallel algorithms for training GNNs and show that they can asymptotically reduce communication compared to previous parallel GNN training methods. We implement these algorithms, which are based on 1D, 1.5D, 2D, and 3D sparse-dense matrix multiplication, using torch.distributed on GPU-equipped clusters. Our algorithms optimize communication across the full GNN training pipeline. We train GNNs on over a hundred GPUs on multiple datasets, including a protein network with over a billion edges. 
\end{abstract}


\section{Introduction}
Graph Neural Networks (GNNs)~\cite{scarselli2008graph} are types of neural networks that use the connectivity information that is natural in datasets that can be represented as graphs, such as 
molecules, transportation and social networks, the power grid, and proteins. The neighborhood connectivity information in GNNs is unrestricted and potentially irregular giving them greater applicability than convolutional neural networks (CNNs), which impose a fixed regular neighborhood structure. 
GNNs have been successful in many application domains and often take advantage of specialized variations such as recurrent GNNs, spatial-temporal GNNs, graph convolutional networks, and graph autoencoders. High-quality surveys of GNNs describe these variations and their applications in more detail~\cite{wu2020comprehensive, zhou2018graph}. GNNs are also provably quite powerful and, for example, are known  to be equivalent to the powerful Weisfeiler-Lehman algorithm for graph isomorphism when the GNNs' underlying aggregation and combination operators are sufficiently flexible~\cite{xu2018how}. 

Two dominant uses of GNNs are {\em node embedding}, which predicts certain properties of individual vertices in a large graph, and {\em graph embedding}, which predicts certain properties of whole graphs. The presentation of this work follows the node embedding case but the general techniques we introduce are applicable to the graph embedding case as well. 

As with CNNs, the training algorithms for GNNs are based  on variations of gradient descent, and typically use the idea of {\em mini-batching} in which a small set of training samples are  evaluated and then used to update the model in a single step. Mini-batching has two purposes.  First, it allows the neural neural to train on a smaller memory footprint. Second, it strikes a good balance between achieving high-performance through higher arithmetic intensity and achieving good convergence. Unlike images in a database, vertices of a graph are dependent on each other, which is one reason behind GNNs expressiveness. However in the case of GNNs, this dependency makes it hard to process a mini-batch of vertices. After only a few layers, the chosen mini-batch ends up being dependent on the whole graph. This phenomenon, known as the {\em neighborhood explosion}, completely nullifies the memory reduction goals.  

To overcome {\em neighborhood explosion}, researchers resort to sophisticated sampling-based algorithms that can help GNN training have a smaller memory footprint by reducing the number of $k$-hop neighbors considered. Sampling algorithms, however, come with approximation errors. Here, we use the aggregate memory of a cluster or supercomputer to train GNNs without mini-batching, similar to other work that use distributed memory to train GNNs~\cite{zhu2019aligraph, mlsys2020_83}. In particular, ROC~\cite{mlsys2020_83} showed that (1) full gradient descent can be competitive with mini-batching in terms of performance, and (2) sampling based methods can lead to lower accuracy. We build on this work by presenting distributed  algorithms with  reduced communication. Our distributed algorithms are general and while presented for full gradient descent, they can be easily modified to operate on a mini-batch setting. 

Training of GNNs can be memory limited on single node machines, and we support training on distributed-memory architectures where two primary challenges are communication costs and load balance. This paper primarily focuses on minimizing communication costs for GNN training. Some of the best algorithms presented in this paper, namely the 2D and 3D algorithms, also automatically address load balance through a combination of random vertex permutations and the implicit partitioning of the adjacencies of high-degree vertices. 

The primary contribution of our paper is the presentation of parallel GNN training algorithms that reduce communication, which are fundamentally different than existing approaches for GNN training. On $P$ processes, our 2D algorithm, which consumes optimal memory, communicate a factor of $O(\sqrt{P})$ fewer words than commonly utilized vertex-partitioning based approaches. The 3D algorithm we describe reduces the number of words communicated by another factor of $O(P^{1/6})$ at the expense of higher memory consumption. \revisiontwo{Finally, our 1.5D algorithm optimizes communication for a given memory footprint, reducing communication volume by a factor of $O(c)$ and latency cost by a factor of $O(c^2)$ at the expense of asymptotically increasing memory footprint by $O(c)$.} Our work presents  algorithmic recipes to get the fastest GNN implementations at large scale.

Our distributed algorithms can be implemented in any system that allows arbitrary divisions of tensors to processes, such as Mesh-Tensorflow~\cite{shazeer2018mesh}. We opted to use PyTorch for our demonstration due to its ubiquity and excellent support for existing GNN models through PyTorch Geometric. All of our experiments are run on the Summit supercomputer at the Oak Ridge Leadership Computing Facility (OLCF). We used the existing single-node kernel implementations in cuSPARSE that are easily called from PyTorch. Faster implementations of key kernels  such as sparse matrix times tall-skinny dense matrix (SpMM) exist~\cite{yang2018design, ipdps14} and would decrease our overall runtime. Faster single node kernels are equivalent from a relative cost perspective to running on clusters with slower networks; both would make our reduced-communication algorithms more beneficial. 

Our current implementations operate on the standard real field, but they can be trivially extended to support arbitrary {\tt aggregate} operations to increase the expressive power of GNNs~\cite{xu2018how}. For example, many distributed libraries such as Cyclops Tensor Framework~\cite{solomonik2013cyclops} and Combinatorial BLAS~\cite{bulucc2011combinatorial} allow the user to overload scalar addition operations through their semiring interface, which is exactly the neighborhood {\tt aggregate} function when applied to graphs. 
Finally, we note that while our focus is on GNN training, all of our algorithms are applicable to GNN inference.
All of our code is available publicly as the CAGNET (Communication-Avoiding Graph Neural nETwork) package at \href{https://github.com/PASSIONLab/gnn_training}{\color{blue}{https://github.com/PASSIONLab/CAGNET}}.

\section{Related Work}
Parallelism opportunities in the training of deep neural networks (DNNs) have been studied intensively in the recent years~\cite{ben2019demystifying}. For DNNs generally, the two broad cases of parallelism are model and data parallelism. Data parallelism replicates the DNN model in the memory of each process and only partitions the data. Data parallelism can be sub-classified into sample (also called batch) and domain  parallelism. In the particular case of convolutional neural networks (CNNs), domain parallelism is often referred to as spatial parallelism~\cite{dryden2019improving}. Model parallelism, on the other hand, partitions the model explicitly. In the common case, each DNN layer can be partitioned into all processes and layers can be computed in their original order. Alternatively, inter-layer pipeline parallelism can be exploited for certain parts of the computation, but not all. For CNNs, further dimensions of parallelism in the form of filter and channel are exploitable as special cases of model parallelism~\cite{dryden2019channel}.  

This might seem like a daunting list of parallelism opportunities to consider for GNN training. Fortunately, as shown in the next section, GNN training is simply a series of algebraic transformations on sparse and dense matrices. Consequently, one can achieve highly-parallel algorithms without even considering the semantic meaning of the dimensions that are partitioned by the algorithm. Ours is similar to an approach taken in earlier work in parallelizing the training of fully-connected and convolutional neural networks~\cite{gholami2017integrated}. Differently for GNNs, the issues of sparsity and load balance play a prominent role in performance and scalability. Our paper provides several different classes of algorithms that take advantage of many different parallelism opportunities available. We asymptotically analyze the communication costs of the algorithms we present, as well as potential alternatives. 

Existing work in parallel GNN training implement their algorithms in specialized frameworks~\cite{mlsys2020_83,neugraph,zhu2019aligraph}. This requires practitioners to port their models and code to that framework, which might be impossible given the lack of an ecosystem to rapidly implement different GNN algorithms. We implement our algorithms using Pytorch~\cite{paszke2019pytorch}, utilizing  torch.distributed and PyTorch Geometric libraries.
Given the wide availability and popularity of PyTorch, not to mention the vast set of GNN variants implemented in PyTorch Geometric~\cite{Fey/Lenssen/2019}, any practitioner with access to a distributed cluster can easily utilize our algorithms to scale their models.

The other PyTorch based distributed graph embedding libraries we are aware of are PyTorch-BigGraph (PBG)~\cite{lerer2019pytorchbiggraph} and Deep Graph Library (DGL)~\cite{dgl}. PBG's website explicitly says that it is not for use with models such as graph convolutional networks and deep networks. Consequently, while it presents some interesting ideas, PBG does not seem to have the expressiveness required to implement GNNs. By contrast, our distributed algorithms can be used to implement anything that is supported by PyTorch Geometric, which already implements a vast majority of top GNN models in the literature.
On the other hand, DGL is an active large-scale project that provides a convenient graph-based interface instead of exposing sparse matrices to the user, and it automatically fuses operations to avoid unnecessary data movement and computation. Our algorithmic work is complementary and can incorporated into DGL in the future. 

The details of data partitioning in various GNN training systems are light. ROC~\cite{mlsys2020_83} advocates a specialized graph partitioning method, and shows that it scales better than random vertex and edge partitioning. AliGraph~\cite{zhu2019aligraph} mentions that it implements both a graph partitioning based approach and a 2D partitioning approach, but does not give any details or provide communication cost analyses. 

\section{Background}

\subsection{Notation}

Table~\ref{symboltable} summarizes the notation used in our paper. \revisiontwo{There is a unique sparse matrix $\mA$ that represents the graph structure but there are $L$ distinct $\mH$ and $\mG$ matrices, indexed $l=0\ldots L - 1$, which are embedding matrices and their derivatives, respectively. Finally there are $L-1$ weight matrices $\mW$ and $\mY$, indexed $l=0\ldots L - 2$, because the number of transitions between feature vectors are one less than the number of embedding matrices.}

When analyzing communication costs we use the $\alpha-\beta$ model where each message takes a constant $\alpha$ time units latency regardless of its size plus an inverse bandwidth term that takes $\beta$ time units per word in the message, to reach its destination. Thus, sending a message of $k$ words takes $\alpha + \beta k$ time.
In addition, we use $\dnnz(\mA)$ when referring to the number of nonzeros in the sparse adjacency matrix $\mA$, which is equal to the number of edges in the graph with self loops added. We also use $d$ for the average degree of a vertex in $\mA$, i.e. $\dnnz(\mA) = dn$.

\begin{table}[t]
\centering
\footnotesize
% \textbf{Table 2: List of symbols and notations used by our algorithm}
\caption{List of symbols and notations used by our algorithm} \label{symboltable}
\begin{tabular}{ |p{2cm}||p{5.5cm}|}
\hline
\multicolumn{2}{|c|}{Symbols and Notations} \\
\hline
Symbol & Description  \\
\hline
$\mA$ & Modified adjacency matrix of graph ($n \times n$)\\
$\mH^l$ & Embedding matrix in layer $l$ ($n \times f$)\\
$\mW^l$ & Weight matrix in layer $l$ ($f \times f$)\\
$\mY^l$ & Matrix form of $\frac{\partial \mathcal{L}}{\partial W^l_{ij}}$ ($f \times f$)\\
$\mZ^l$ & Input matrix to activation function ($n \times f$)\\
$\mG^l$ & Matrix form of $\frac{\partial \mathcal{L}}{\partial Z^l_{ij}}$ ($n \times f$)\\
$\sigma$ & Activation function\\
$f$ & Length of feature vector per vertex \\
$f_u$ & Feature vector for vertex $u$   \\
$L$ & Total layers in GNN \\
$P$ & Total number of processes \\
$\alpha$ & Latency \\
$\beta$ & Reciprocal bandwidth \\
\hline 
\end{tabular}
\end{table}

\hide{
\begin{table}[!h]
\centering
\caption{Matrices in an $L$-layer GNN}
\begin{tabular}{ |p{2cm}||p{3cm}|}
\hline
Matrix & Layers (i.e. values of $l$)  \\
\hline
$\mA$ & N/A\\
$\mH^l$ & $0\ldots L - 1$\\
$\mG^l$ & $0\ldots L - 1$\\
$\mW^l$ & $0\ldots L - 2$\\
$\mY^l$ & $0\ldots L - 2$\\
\hline
\end{tabular}
\end{table}
}

\subsection{Graph Neural Networks}

Consider a dataset that is represented as a graph $G(V,E)$, such as a protein network, social network, grid transmission network, or the union of tangled high-energy particle tracks. Here, $V$ is the set of vertices (nodes) and $E$ is the set of edges. We can consider the classification of the nodes or the edges. Without loss of generality, we will describe a GNN for node classification. The goal of the so-called {\em node embedding} problem is to map the nodes of a graph into a low-dimensional embedding space such that the similarity of two nodes $u, v \in V$ is approximated by their similarity in the low-dimensional space $z_u\transpose z_v$. Here, $z_v$, which is typically a $k$-dimensional vector of floating-point values, is the embedding of vertex $v$. In addition to node and edge classification, GNNs can also be used to classify graphs or perform regression on graphs. In this case, the input would be a set of graphs such as the atomistic structures of a set of molecules. 

Let $\mA$ be the $n \times n$ sparse adjacency matrix of the graph with added self-connections. The addition of self-connections ensures that each node does not forget its embedding when going from layer $i$ to layer $i+1$. The rows and columns of $\mA$ are also often normalized~\cite{KipfWelling2017}, so for an undirected graph one actually uses $\mathbf{D}^{-1/2}(\mA+\mathbf{I})\mathbf{D}^{-1/2}$ due to its favorable spectral properties. Here, $\mathbf{I}$ is the identity matrix and $\mathbf{D}$ is a diagonal matrix of modified vertex degrees. To avoid notational burden, we will still refer to this modified adjacency matrix with $\mA$. We also distinguish between $\mA$ and its transpose $\mA\transpose$ explicitly in order to present a general training algorithm that works for both directed and undirected graphs. $\mH^0$ is a dense $n \times d$ matrix of input node features. These features are application dependent attributes on graph nodes. A high-quality embedding can be achieved by using a neural network that uses the topology of the graph. In particular, the GNN forward propagation processes t
he input features matrix $\mH^{(l)}$ at level $l$ using following simple equation:
$\mH^{(l)}= \sigma ( \mA\transpose \mH^{(l-1)} \mW^l ).$

Here, $\mW^{(l)}$ is the {\em trainable matrix} that holds the model parameters at the $l$th level of the neural network, and $\sigma$ is the activation function such as $\textrm{ReLU}$. Consequently, the most time consuming operations are the multiplication of a sparse matrix with a dense matrix (SpMM) and dense matrix multiply. Backpropagation also relies on the same computational primitives. We provide backpropagation derivations in Section~\ref{sec:backpropagation}. 

\subsection{Forward Propagation}
\revisiontwo{At each node, the product $\mA\transpose \mH^{(l - 1)}$ combines the $(i-1)$th feature vectors of its neighbors while the subsequent multiplication with $\mW^l$ mixes the features and maps them into the new feature space at the $i$th level. Finally, nonlinearity is achieved via the $\sigma()$ function on the output.}
\begin{align*}
    \mZ^l &= \mA\transpose \mH^{(l - 1)}\mW^l \\ 
    \mH^l &= \sigma (\mZ^l)
\end{align*}

\subsection{Backpropagation Derivations}
\label{sec:backpropagation}
\subsubsection*{Equation 1}
\revision{Here we derive the gradient of the loss with respect to $\mZ^L$, leveraging the chain rule and that $\mZ^l = \sigma(H^l)$.}
    \begin{align*} 
        \frac{\partial \mathcal{L}}{\partial Z^L_{ij}} &= \sum_{u \in V}\sum_{v \in f_u}\frac{\partial \mathcal{L}}{\partial H^L_{uv}}\frac{\partial H^L_{uv}}{\partial Z^L_{ij}} \\ 
        &= \frac{\partial \mathcal{L}}{\partial H^L_{ij}} \frac{\partial H^L_{ij}}{\partial Z^L_{ij}} \tag{$\frac{\partial H^L_{uv}}{\partial Z^L_{ij}} = 0$ iff $u \neq i$ and $v \neq j$} \\
        &= \frac{\partial \mathcal{L}}{\partial H^L_{ij}} \sigma '(Z^L_{ij}) \\
    &\boxed{G^L = \frac{\partial \mathcal{L}}{\partial Z^L_{ij}} = \nabla_{H^L}\mathcal{L} \odot \sigma '(Z^L)}
    \end{align*}

\subsubsection*{Equation 2}
\revision{Here we derive a recurrence to propagate the gradient backwards through the neural network, leveraging the chain rule and the lemma stated below.}
    \begin{align*}
        G^{l - 1}_{ij} = \frac{\partial \mathcal{L}}{\partial Z^{l - 1}_{ij}} &= \sum_{u \in V}\sum_{v \in f_u}\frac{\partial \mathcal{L}}{\partial Z^{l}_{uv}} \frac{\partial Z^{l}_{uv}}{\partial Z^{l - 1}_{ij}} \\
        &= \sum_{u \in V}\sum_{v \in f_u}\frac{\partial Z^{l}_{uv}}{\partial Z^{l - 1}_{ij}}G^{l}_{uv} \\
        &= \sum_{u \in V}\sum_{v \in f_u}W^{l}_{jv}G^{l}_{uv}\sigma '(Z^{l - 1}_{ij}) \tag{see lemma below} \\
        &\boxed{\mG^{l - 1} = \mA \mG^{l}(\mW^{l})\transpose \odot \sigma '(\mZ^{l - 1})}
    \end{align*}

{\bf Lemma for Equation 2:}
    \begin{align*}
        Z^{l}_{uv} &= \sum_{i \in N(u)}\sum_{j \in f_i}H^{l - 1}_{ij}W^{l}_{jv} \tag{See forward prop equations} \\        
        &= \sum_{i \in N(u)}\sum_{j \in f_i}\sigma(Z^{l - 1}_{ij})W^{l}_{jv} \\        
        \frac{\partial Z^{l}_{uv}}{\partial Z^{l- 1}_{ij}} &= W^{l}_{jv}\sigma '(Z^{l - 1}_{ij})
    \end{align*}

\subsubsection*{Equation 3}
\revision{This final equation represents the gradient $\mY^l$ of the loss with respect to the weights in the network, and this gradient is used in gradient descent.}
%\begin{center}
    \begin{align*}
        \mY^{l - 1} = \Bigg(\frac{\partial \mathcal{L}}{\partial W^l}\Bigg)_{ij} &= (\mH^{l - 1})\transpose \mA\mG^l \\
        \mW^{l - 1} &= \mW^{l - 1} - \mY^{l - 1}
    \end{align*}
%\end{center}
\revision{The second step in Equation $3$ is simply the gradient descent step. This step does not require communication, so it is not discussed in our analysis in the following section.}

\section{Communication Schemes and their Analyses}
In this section, we present 1D, 1.5D, 2D, and 3D parallel algorithms for GNN training and analyze their communication costs. \revisiontwo{Table~\ref{table:distribution} summarizes the matrix partitioning used by these algorithms}.
The presented communication costs are for one epoch, which is a single pass over the whole dataset.

\revisiontwo{Ideally, a distributed-memory parallel GNN training algorithm consumes $O(nfL + \dnnz(\mA))$ total memory across all processes. Our 1D and 2D algorithms achieve this bound up to constant factors, while 1.5D and 3D algorithms, whose exact memory consumption is provided in their respective subsections, do not. 
To avoid expensive transposing, our 1D and 1.5D algorithms stores both $\mA$ and $\mA\transpose$ when the input graph is directed. This decision does not increase storage costs asymptotically and is currently immaterial due to almost all GNN-based learning being performed on undirected graphs.}

\revisiontwo{All pseudocodes (Algorithms~\ref{alg:blockrow1Dforward},\ref{alg:blockrow15dforward},\ref{alg:sparsesumma}) take the inputs 
\begin{enumerate}
    \item $\mA \in \mathbb{R}^{n \times n}:$ sparse adjacency matrix,
    \item $\mH^{l-1} \in \mathbb{R}^{n \times f^{l-1}}:$ dense input activations matrix,
    \item $\mW \in \mathbb{R}^{f^{l-1} \times f^{l}}:$ dense training matrix,
\end{enumerate} and output $\mH^l: \mathbb{R}^{n \times f^l}:$ dense output activations matrix.}

\begin{table*}[!h]
\label{table:distribution}
\centering
\caption{\revisiontwo{Data distribution for all the algorithms considered in our paper}}
\begin{tabular}{ |p{1cm}||p{2cm}||p{4cm}||p{2.5cm}||p{2.5cm}|}
\hline
Matrix & 1D Algorithm & \revisiontwo{1.5D Algorithm} & 2D Algorithm & 3D Algorithm \\
\hline
$\mA$ & Block row & \revisiontwo{Block row, replicated $c$ times} & Block 2D & Block Split 3D \\
$\mA\transpose$ & Block row & \revisiontwo{Block row, replicated $c$ times} & Not stored & Not stored \\
$\mH^l$ & Block row & \revisiontwo{Block row, replicated $c$ times} & Block 2D & Block Split 3D \\
$\mG^l$ & Block row & \revisiontwo{Block row, replicated $c$ times} & Block 2D & Block Split 3D \\
$\mW^l$ & Fully-replicated & \revisiontwo{Fully-replicated} & Fully-replicated &  Fully-replicated \\
\hline
\end{tabular}
\end{table*}

\subsection{A One-Dimensional (1D) Algorithm}

\hide{
\begin{table}[!h]
\centering
\caption{Data distribution for our 1D algorithm}
\begin{tabular}{ |p{2cm}||p{3cm}|}
\hline
Matrix & Partitioning  \\
\hline
$\mA$ & Column-based\\
$\mH^l$ & Row-based \\
$\mG^l$ & Row-based \\
$\mW^l$ & Fully-replicated \\
\hline
\end{tabular}
\end{table}}

In this regime, matrices $\mA\transpose$ and $\mH$ are distributed to processes in block rows, where each process receives $n/P$ consecutive rows. 
For example, given a matrix $\mA\transpose$, we write \revision{$\mA\transpose_i = \mA\transpose(i (n/P):(i+1)(n/P)-1,:)$} to denote the block row owned by the $i$th process, assuming $n/P$ is an integer. To simplify the algorithm description, 
we use $\mA\transpose_{ij}$ to denote \revision{$\mA\transpose_i (:, j (n/P):(j+1)(n/P)-1)$}, the $j$th block column of $\mA\transpose_i$, although the whole block row is owned by a single process.

\begin{equation}
\mA\transpose = \left( 
\begin{array}{c}
\mA\transpose_{1} \\
%\hline
\vdots \\
%\hline
\mA\transpose_{p}
\end{array} 
\right)
= \left( 
\begin{array}{c c c}
\mA\transpose_{11} & \ldots  & \mA\transpose_{1 p} \\
%\hline
\vdots  & \ddots  & \vdots  \\
%\hline
\mA\transpose_{p 1} & \ldots   & \mA\transpose_{p p} 
\end{array} 
\right), 
\mH = \left( 
\begin{array}{c}
\mH_{1} \\
%\hline
\vdots \\
%\hline
\mH_{p}
\end{array} 
\right)
\label{eqn:1dpartitioning}
\end{equation}

Let $\mT$ be the intermediate product $\mA\transpose \mH^{l - 1}$. For each process $P(i)$, the computation is:
$$ \mT_i = \mT_i + \mA\transpose_i \, \mH = \mT_i + \sum_{j=1}^{p} \mA\transpose_{ij}  \, \mH_j $$

The row-wise algorithm forms one row of output at a time, and each process may potentially need to access all of $\mH$ to form a single row of $\mT$. 
However, only a portion of $\mH$ is locally available at any time in parallel algorithms. The algorithm, thus, performs multiple iterations to fully 
form one row of $\mT$. Algorithm~\ref{alg:blockrow1Dforward} shows the pseudocode of 
the algorithm. 

\begin{algorithm}
\begin{algorithmic}[1]
%\Require $\mA \in \mathbb{R}^{n \times n}:$ sparse adjacency matrix, 
%$\mH^l \in \mathbb{R}^{n \times f^{l-1}}:$ dense input activations matrix, 
%$\mW \in \mathbb{R}^{f^{l-1} \times f^{l}}:$ dense training matrix 
%\Ensure $\mH^l \in \mathbb{R}^{n \times f^l}:$ dense matrix of layer $l$ activations 
\Procedure{BlockRowFW}{$\mA\transpose, \mH^{l-1}, \mW, \mH^l$}
\For{all processes $P(i)$\  \InParallel} 
\For{ $j =1$ to $p$}
\State \Call{Broadcast}{$\mH^{l-1}_{j}$} 
\State $\mT_{ij} \gets \mA\transpose_{ij} \mH^{l-1}_{j}$ 
\State $\mZ_{i} \gets \mT_{ij} \mW$
\State  $\mH^{l}_{i} \gets \mH_{i}^{l} + \sigma(\mZ_i)$ 
\EndFor
\EndFor
\EndProcedure
\end{algorithmic}
\caption{Parallel algorithm for GNN forward propagation, which computes $\mH^{l} \gets \sigma(\mA\transpose \mH^{l - 1}\mW^l)$, using the 1D block row decomposition}\label{alg:blockrow1Dforward} 
\end{algorithm}

\subsubsection{Equation $\mZ^l = \mA\transpose \mH^{l - 1}\mW^l$}
%\begin{figure}[h]
%    \centering
%    \includegraphics[scale=0.4]{ps1eq1_new.png}
%    \caption{Partitioning Scheme $1$ Equation $1$}
%    \label{fig:ps1eq1}
%\end{figure}

\paragraph*{Communication is 1D Block Row} $\mA\transpose$ is partitioned by rows, and $\mH^{l - 1}$ is partitioned by rows. This yields a $1$D Block Row multiplication. $\mW^l$ is fully-replicated on each process, and is multiplied with $\mA\transpose \mH^{l - 1}$ after communication.  The first multiplication is essentially a sparse matrix times (tall-skinny) dense matrix, also known as sparse matrix times multiple dense vectors (SpMM). 

\hide{In the context of distributed SpMM, we use $\mathit{edgecut}_P(\mA)$ to refer to $\max(r_1, \ldots, r_P)$, where $r_i$ is the minimum number of dense matrix rows process $i$ needs to communicate to any other process in order for it to perform  local matrix multiplication. This cost needs to be multiplied with $f$ because each such row is of length $f$, i.e. each node carries a payload of size $f$ in the form of a feature vector. This is illustrated in Figure~\ref{fig:graphcut}.}

\hide{Graph and hypergraph partitioning tools can be applied as pre-processing to heuristically minimize the $\mathit{edgecut}_P(\mA)$ metric~\cite{bulucc2016recent}. Note that a non-adversarial $\mathit{edgecut}_P(\mA)$ is never higher than $n(P-1)/P$, which can be achieved by a random partitioning. Nevertheless, we chose to present the bounds in terms of $\mathit{edgecut}_P(\mA)$ because it can be lower than $n(P-1)/P$ thanks to aforementioned tools.}

\revisiontwo{
Our 1D algorithm moves the dense matrix in this SpMM operation using a broadcast. The alternative approach of moving the sparse matrix would yield a similar communication cost in practice because the dense feature matrices in GNNs have approximately the same size (in terms of bytes) as the graphs they are run on. As input and model trends change in the future, a simple heuristic can determine the matrix to be broadcasted in practice, without increasing code complexity.}

\revisiontwo{Alternatively, one could shift matrices as opposed to broadcasting them. Point-to-point communication is still in beta in NCCL, the library we use for communication. The cost of a single broadcast of an $m$ word message to $P$ processes has a lower bound of $O(\alpha\lg{P}+\beta m)$~\cite{chan2007collective}, but high-level algorithms such as SUMMA~\cite{van1997summa} can avoid the $\lg{P}$ factor in the latency term through pipelining. In fact, NCCL uses a pipelined ring algorithm for its broadcasts, which in fact achieves the same link utilization as matrix shifting when the message sizes are large enough to fill the pipeline. Consequently, we will also not spell out the additional $\lg{P}$ factor for our  broadcasts.}  

The per-process communication cost is thus 
$$T_{comm} = \alpha (P-1) + \frac{P-1}{P} \beta \, n  f \approx  \alpha P + \beta \, n f$$
    
\hide{
\begin{figure}
\begin{center}
\includegraphics[width=\columnwidth]{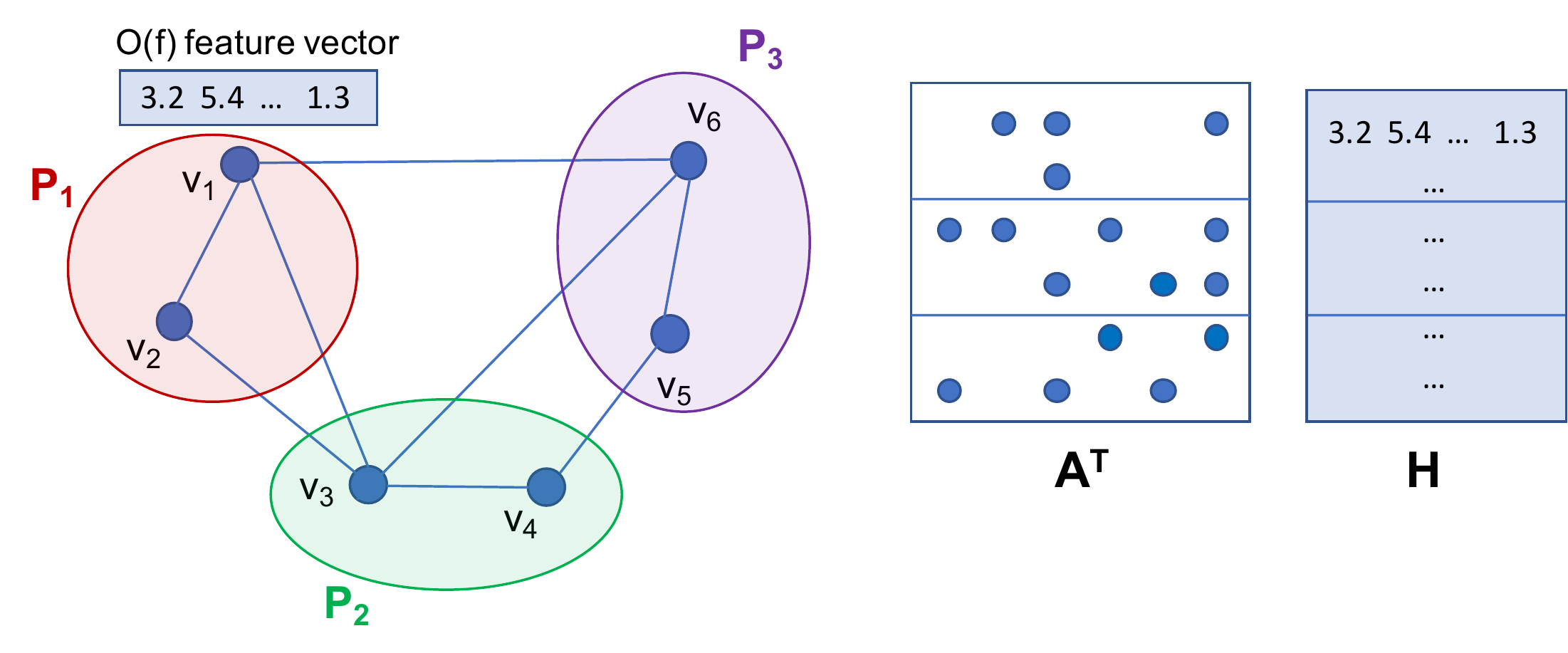}
\vspace{-0.75cm}
\caption{Illustration of the graph edge cut cost for partitioning scheme 1. For example, since $v_1$ has a connection to $v_6$, $P_1$ needs to receive $v_6$'s $O(f)$ feature vector from $P_3$. This is modeled by the edge $(v_1,v_6)$ that is ``cut'' between those two partitions. }
\vspace{-0.5cm}
\label{fig:graphcut}
\end{center}
\end{figure}}

\subsubsection{Equation $\mH^l = \sigma(\mZ^l)$}
\paragraph*{No Communication} 
    $\mH^l$, the result of activation, is partitioned by rows as is $\mH^{l - 1}$. No further communication is necessary here to use $\mH^l$ in Eq. 1 for layer $l$.
    
\subsubsection{Equation $\mG^{l - 1} = \mA \mG^{l}(\mW^{l})\transpose \odot \sigma '(\mZ^{l - 1})$}
%\begin{figure}[h]
%    \centering
%    \includegraphics[width=\columnwidth]{ps1eq2_new.png}
%    \caption{Partitioning Scheme $1$ Equation $2$}
%    \label{fig:ps1eq2}
%\end{figure}

\paragraph*{Communication is 1D Block Row} 
\revisiontwo{
Because we also partition $\mA$ in block rows, the communication pattern and the cost is identical to the forward propagation. The intermediate product $\mA\mG^{l}$ is naturally block row partitioned. The last step of multiplying the block row distributed $\mA\mG^{l}$ with replicated $\mW^{l}$ to yield a block row distributed $\mG^{l-1}$ does not require any communication.}

\hide{$\mA$ is partitioned by columns, and $\mG^{l}$ is partitioned by rows. This yields a $1$D Outer Product multiplication. The communication results from the reduction of low-rank matrices, each of which are of size $n \times f$, formed on each process. The result of this reduction is $\mG^{l-1}$, which is partitioned by rows just like $\mG^{l}$ so that the next iteration of the backpropagation does not incur a redistribution cost. }
    
\hide{The intermediate low-rank products $\mA_{i} \mG^{l}_i$ are dense unless there are empty rows on each part of $\mA$. For the sake of initial analysis, let us assume we have an Erd\H{o}s-R\'enyi graph $G(n, d/n)$ where each possible directed edge occurs with probability $d/n$. This results in $n^2  (d/n) \approx dn$ edges.  Following the analysis in Section 4.1.2 of earlier work~\cite{ballard2013communication}, the expected number of non-empty rows on each part $\mA_i$ is $dn/P$, for large $P$ (i.e. $P>d$).  The storage cost of the intermediate low-rank product $\mA_i  \mG_i^{l}$ is expected to be $O(dnf/P)$, which is high but memory-scalable with increasing process counts. At large scale (i.e. when $P> d$), this is better than storing the intermediate products in dense format, which would require a storage cost of $O(nf)$ per process. 
    
 The theoretical sparsity analysis in the previous paragraph makes a case for taking advantage of sparsity for intermediate low-rank products for large $P$. The practical issue is that sparse reduction algorithms carry high overheads that makes this approach hard to optimize, though there have been several encouraging developments in this area~\cite{renggli2019sparcml}. }
     
  %  We do not have to perform the expensive reduction at this point because that can be deferred until after the next multiplication with $(\mW^{l})\transpose$, which is fully-replicated on each process due to its small size. This is because matrix multiplication
  %  is distributive over matrix addition, yielding
  %  \begin{align*}
  %  \mA \mG^{l} (\mW^{l})\transpose  &= (\sum_{i=1}^{p} \mA_i \mG^{l}_i)  (\mW^{l })\transpose  \\
    %							&= \sum_{i=1}^{p} (\mA_i \mG^{l}_i  (\mW^{l})\transpose).
%							    \end{align*}
							    
\hide{At this point, each process has $O(nf)$ data that need to be reduce-scattered in order to get the intermediate product $\mA\mG^{l}$ in a block row-distributed state. The communication cost is 
$T_{comm} = \alpha \lg{P} + \beta nf (P-1)/P $~\cite{chan2007collective, thakur2005optimization}, which we round up to 
$T_{comm} = \alpha \lg{P} + \beta nf $ to reduce clutter. }

\subsubsection{Equation $\mY^{l-1}= (\mH^{l - 1})\transpose \mA \mG^l$}
%\begin{figure}[h]
%    \centering
%    \includegraphics[width=\columnwidth]{ps1eq3_new.png}
%    \caption{Partitioning Scheme $1$ Equation $3$}
%    \label{fig:ps1eq3}
%\end{figure}

\paragraph*{Communication is (small) 1D Outer Product} 
    Algebraically, there are two matrix multiplications in this step of the backpropagation. \revision{However, we can reuse the intermediate product $\mA\mG^{l}$ that we computed in the previous equation at the expense of increasing the memory footprint slightly.} Then the only task is to multiply $(\mH^{l - 1})\transpose$ and $\mA \mG^{l}$, which is a small 1D outer product that requires an all-reduce on low-rank matrices of size $f \times f$. This multiplication has communication cost $T_{comm} = \alpha \lg{P} + \beta \, f^2$.

    %which is the column-wise analog of the 1D block row multiplication described for the forward step, resulting in the same communication cost $T_{comm} = \alpha P + \beta \, \mathit{edgecut}_P(\mA) \, f$.  
    
    %The second is multiplying the result of the first product with $\mG^l$,  
    
\subsubsection{Total Communication of our 1D Algorithm}
Given that the embedding (i.e., feature vector) lengths are different for each layer of the GNN, we use the superscript to denote the length of the feature vector $f^l$ in layer $l$. This results in the following
communication bound.
\begin{align*}
    T_{comm} = \sum_{l = 1}^{L}\Bigl(\alpha(\lg{P} + 2P) + \beta \bigl(nf^{l-1} + nf^l + f^{l-1} f^l )\bigr) \Bigr)
\end{align*}

To reduce clutter, we can consider the ``average" feature vector length $f$, resulting in the simplified formula.
\begin{align*}
    T_{comm} = L\, \bigl(\alpha(\lg{P} + 2P) + \beta( 2nf + f^2) \bigr)
\end{align*}

\hide{
\subsubsection{The symmetric case}
The ability to treat $\mA$ as if it is $\mA\transpose$ and vice-versa gives the algorithm the freedom to trade 1D outer products with 1D block row/column multiplications. Given than $\mathit{edgecut}_P(\mA) \leq n(P-1)/P$, the resulting (simplified) communication cost for the symmetric case is  
\begin{align}
T_{comm} = L\, \bigl(\alpha(3\lg{P}) + \beta(2\, \mathit{edgecut}_P(\mA) \, f + f^2) \bigr). 
\label{eq:sym1d}
\end{align}}

\hide{
\subsubsection{Discussion of 1D alternatives}
There are several other 1D algorithms we considered. For example, we could partition $\mA$ row-wise as opposed to column-wise. This makes the forward propagation an outer product instead of a 1D block row. Conversely, the first step of the backpropagation, $\mG^{l - 1} = \mA \mG^{l}(\mW^{l})\transpose$, becomes a 1D block row as opposed an outer product. 
The last step of the backpropagation, $\mY^{l-1}= (\mH^{l - 1})\transpose \mA \mG^l$, still costs the same in terms of communication as it performs a small 1D outer product. Consequently, we would still be performing 1 large outer product, 1 small outer product,  and 1 block row multiplication as before, resulting in the same total communication cost. 

We also considered transposing $\mA$ between forward and backward propagation, ensuring that both $\mA$ and $\mA\transpose$ are partitioned the same way (either by rows or columns). If the input is undirected, this is already the case without additional transpositions and degenerates to the symmetric case presented above. \todo{Make this the default case: Similarly, if one can afford store two copies of adjacency matrices, then the symmetric case bound in Equation~\ref{eq:sym1d} applies.} The communication cost of transposition is $\alpha p^2 + \beta \dnnz(\mA)/P$ and has to be done only twice per epoch (once after forward propagation and  once after backpropagation), not at every layer. The communication costs of this transposing variant would therefore be
\begin{align*}
T_{comm} &= 2 \alpha p^2 + 2 \beta \dnnz(\mA)/P  \\
	&+ L\, \bigl(\alpha(3\lg{P}) + \beta(2\, \mathit{edgecut}_P(\mA) \, f + f^2) \bigr). 
\end{align*}}

\subsubsection{Potential for graph partitioning}
\revisiontwo{Since we are moving dense matrices and keeping the sparse matrix stationary, graph and hypergraph partitioning tools can be applied as pre-processing to heuristically minimize communication.} We experimented with graph partitioning to evaluate its potential for us. We ran Metis on the Reddit data, \revision{which is described in Section~\ref{sec:data}}. For 64 processes, Metis' partitions suggested a $72\%$ total communication reduction over random block row distribution.
%($\mathit{edgecut}$ of $3258385$ vs $11761151$). 
However, the total runtime of our bulk-synchronous algorithm would be dictated by the maximum communication per process, which was only  a $29\%$ percent reduction over random 1D block row partitioning.
%($131286$ vs $185823$ edges cut for the process with maximum communication). 
These numbers are actually optimistic and do not take into account the need to perform individualized ``request and send'' operations for exploiting the graph partitioning results.

For example, instead of relying on more efficient broadcast operations as done in Algorithm~\ref{alg:blockrow1Dforward}, each process that owns a part of $\mA$ would (depending on its sparsity structure) individually request a subset of rows of $\mH$ from its owner during forward propagation. This increases latency as well as the bandwidth costs due to the communication of request indices, and makes it impossible to benefit from collective communication libraries such as NCCL and gloo. Further, given the scale free nature of most graph datasets, graph partitioning is unlikely to produce an asymptotic improvement despite its added computational costs. 

\subsection{1.5D Block Row Algorithm}
\label{sec:15d}
% 1.5D algorithms~\cite{spdmmm16} are attractive from the perspective of reducing communication when the matrices to be multiplied are very different in size. Here, we use the word `size' to denote the number of nonzeros in a sparse matrix or the product of its dimensions in a dense matrix. One downside of 1.5D algorithms compared to 2D algorithms is their additional memory consumption due to replication. This was less of a concern for dense deep neural network training where either the model or the data was replicated for parallelism. However, for GNN training, memory is at a premium. Hence, 2D algorithms, which do not use any extra memory, are preferable.

% In the case of GNN training, the relative sizes of two input matrices are $\dnnz(\mA)$ and $n \, f$. When the average degree $d = \dnnz(\mA) /n$ of the graph is close to the embedding vector length $f$, then they are of similar size. In practice for most dataset $d = O(f)$, which makes it hard to justify the added memory burden of 1.5D algorithms. 

\revisiontwo{For 1.5D algorithms~\cite{spdmmm16}, processes are organized in a rectangular $P = P / c \times c$ grid. Matrices are, however, partitioned into block rows and columns as done in 1D. The difference between 1D and 1.5D algorithms is that these partitions are now replicated across process rows. For instance, processes across the $i$th process row $P(i, :)$ collectively store the $i$th block row of $\mA\transpose$. Because of this difference, while matrices are partitioned into block rows and columns, there are only $P / c$ such blocks. }

\revisiontwo{
\begin{equation}
\mA\transpose = \left( 
\begin{array}{c}
\mA\transpose_{1} \\
%\hline
\vdots \\
%\hline
\mA\transpose_{p/c}
\end{array} 
\right)
% = \left( 
% \begin{array}{c c c}
% \mA\transpose_{11} & \ldots  & \mA\transpose_{1 p/c} \\
% %\hline
% \vdots  & \ddots  & \vdots  \\
% %\hline
% \mA\transpose_{p/c 1} & \ldots   & \mA\transpose_{p/c p/c} 
% \end{array} 
% \right), 
\mH = \left( 
\begin{array}{c}
\mH_{1} \\
%\hline
\vdots \\
%\hline
\mH_{p/c}
\end{array} 
\right)
\label{eqn:15dpartitioning}
\end{equation}
}

\revisiontwo{Similar to 1D, each submatrix $\mA\transpose_{i}$ is further partitioned in $p/c$ block columns.}

\revisiontwo{Let $\mT$ be the intermediate product of $\mA\transpose\mH^{l - 1}$. Each process row $P(i, :)$ computes the following:}
\revisiontwo{$$ \mT_i = \mT_i + \mA\transpose_i \, \mH = \mT_i + \sum_{j=1}^{p/c} \mA\transpose_{ij}  \, \mH_j $$}
\revisiontwo{However, each process computes a subset of the terms in the above summation. These partial sums are then added within process rows with a reduction on $P(i, :)$. If $q = p / c^2$, then the computation done by process $P(i, j)$ is}
\begin{equation}
\revisiontwo{ \mT_i = \mT_i + \mA\transpose_i \, \mH = \mT_i + \sum_{k=jq}^{(j + 1)q} \mA\transpose_{ik}  \, \mH_k }
\label{eqn:15dsum}
\end{equation}
\revisiontwo{These steps are outlined in detail in Algorithm~\ref{alg:blockrow15dforward}. While our pseudocode only outlines the special case where $c^2$ perfectly divides $p$, our implementation is more general, and assigns more stages to the last process column if necessary.}

\begin{algorithm}
\begin{algorithmic}[1]
%\Require \revisiontwo{$\mA \in \mathbb{R}^{n \times n}:$ sparse adjacency matrix, $\mH^l \in \mathbb{R}^{n \times f^{l-1}}:$ dense input activations matrix, 
%$\mW \in \mathbb{R}^{f^{l-1} \times f^{l}}:$ dense training matrix}
%\Ensure \revisiontwo{$\mH^l \in \mathbb{R}^{n \times f^l}:$ dense output activations matrix}
\Procedure{\revisiontwo{Block1.5DFW}}{\revisiontwo{$\mA\transpose, \mH^{l-1}, \mW, \mH^l$}}
\For{\revisiontwo{all processes $P(i,j)$ \InParallel}} 
    % \State \revisiontwo{$\mA^T = \mA^T$.split$(n / (p / c))$}
    \State $\revisiontwo{s = p / c^2}$  \Comment{number of stages}
    \For{\revisiontwo{$k = 0$ to $s - 1$}}
        % \State \revisiontwo{$q = (i / (p / c^2)) * (p / c^2) + k$}
        \State \revisiontwo{$q = j \, s + k$}
        \State \revisiontwo{$\hat{\mH}^{l-1} \gets $ \Call{BCast}{$\mH^{l - 1}_{qj}$, $P(:,j)$}}
    	\State \revisiontwo{$\mZ^l \gets \mZ^l + \Call{SpMM}{\mA\transpose_{iq},\hat{\mH}^{l - 1}}$}
    \EndFor
    \State \revisiontwo{$\hat{\mZ}^{l} \gets $ \Call{AllReduce}{$\mZ^{l}$, +, $P(i, :)$}}
    \State \revisiontwo{$\hat{\mH}^{l} \gets $ \Call{GEMM}{$\mZ^{l}$, $\mW^{l - 1}$}}
\EndFor
\EndProcedure
\end{algorithmic}
\caption{\revisiontwo{Block 1.5D algorithm for GNN forward propagation, which computes $\mH^{l} \gets \sigma(\mA\transpose \mH^{l - 1}\mW^l)$ in parallel. $\mA$ and $\mH$ are distributed on a $p / c \times c$ process grid, $\mW$ is replicated.}} \label{alg:blockrow15dforward}
\end{algorithm}

\subsubsection{\revisiontwo{Equation  $\mZ^l = \mA\transpose \mH^{l - 1}\mW^l$}}
\revisiontwo{\paragraph{Communication: 1.5D Block Row.} Both $\mA^T$ and $\mH$ are partitioned by rows in a $P / c \times c$ process grid. We group process rows into $c$ ``chunks'', with $p / c^2$ process rows per chunk. These chunks represent the block rows of $\mH$ that a particular process column accesses, as per Equation~\ref{eqn:15dsum}. To compute a submatrix of $\mA\transpose\mH$, we broadcast each block row to a process column based on its chunk. If a block row is in chunk $i$, we broadcast it to $i$th process column $P(:, i)$. Since there are $p / c^2$ chunks with $c$ blocks row each, each process participates in only $p / c^2$ broadcasts. After these iterations of broadcasts complete, each process within a process row has a partial sum for its submatrix. We run an all-reduction to compute the final block row. Note that $\mW^l$ is fully-replicated, so we do not need to communicate data to multiply with $\mW^l$. The overall communication cost for this equation is $$T_{comm} = \alpha \Big(\frac{P}{c^2}\lg{\frac{P}{c^2}}\Big) + \beta \Big(\frac{nf}{c} + \frac{nfc}{P}\Big)$$}

\subsubsection{\revisiontwo{Equation $\mH^l = \sigma(\mZ^l)$}}
\revisiontwo{\paragraph*{No Communication} 
    $\mH^l$, the result of activation, is partitioned by rows as is $\mH^{l - 1}$. No further communication is necessary here to use $\mH^l$ in Eq. 1 for layer $l$.}
    
\subsubsection{\revisiontwo{Equation $\mG^{l-1} = \mA \mG^{l}(\mW^{l})\transpose \odot \sigma '(\mZ^{l-1})$}}
\paragraph*{\revisiontwo{Communication: 1.5 Block Row}}
    \revisiontwo{Recall that $\mA$ is  partitioned by rows and stored separately from $\mA\transpose$ if graph is directed. $\mG$ is also partitioned by rows. 
    %Ordinarily, this would yield and outer product multiplication. Outer products are not scalable, however, so we opt to transpose the block columns of $\mA$ to make $\mA$ partitioned into block rows. Now 
    Hence, we can apply the same 1.5D algorithm used in Equation 1. We also need to account for $\sigma' (\mZ^{l-1})$. Recall that, as in Equation 2, this step requires no communication as $\mZ^{l - 1}$ is partitioned by rows. The communication cost for this equation is $$T_{comm} = \alpha \Big(\frac{P}{c^2}\lg{\frac{P}{c^2}}\Big) + \beta \Big(\frac{nf}{c} + \frac{nfc}{P}\Big)$$}
    
\subsubsection{\revisiontwo{Equation $\mY= (\mH^{l - 1})\transpose \mA \mG^l$}}
\paragraph*{\revisiontwo{Communication: (small) 1.5D Outer Product}}
    \revisiontwo{We store the intermediate product $\mA \mG^l$ that was computed in the previous step and reuse it here. Multiplying $(\mH^{l - 1})\transpose$ with $\mA \mG^l$ is a dense 1.5D Outer Product on two matrices with $nf$ elements, resulting in a small $f \times f$ output. Because of the small output, this outer product is neither compute nor memory-intensive. The resulting communication cost is: $$T_{comm} = \alpha \Big(\lg\frac{P}{c}\Big) + \beta (f^2)$$}
    
\subsubsection{\revisiontwo{Total Communication}} \revisiontwo{Ignoring $\lg{P}$ latency terms and $f^2$ bandwidth terms, we have a total communication cost of $$T_{comm} = \sum_{l = 1}^{L}\Big(\alpha\Big(2\frac{P}{c^2}\log\frac{P}{c^2}\Big) + \beta\Big(\frac{2nf}{c} + \frac{2nfc}{P}\Big)\Big)$$}

\revisiontwo{While our 1D algorithm did not scale with increasing process counts, we see here that our 1.5D algorithm scales in proportion to the harmonic mean of $P / c$ and $c$. This performance benefit, however, comes at a memory cost. The input for the 1.5D algorithm is replicated, evidenced by multiple processes storing the same data across process rows. Formally, the per-process memory and the total memory are
\begin{align*}
    M^{1.5D}_{proc} = L\Big(\frac{\dnnz(\mA)}{p/c} + \frac{nf}{p/c}\Big) \\
    M^{1.5D}_{total} = Lc(\dnnz(\mA)+ nf)
\end{align*}
}
\revisiontwo{The total input memory used by the 1.5D algorithm is $c$ times more than the input for 1D. This is not negligible, as GNN models tend to be big. In addition, GPUs tend to have less memory, making input replication prohibitive.}
    
%Instead, we opt for a 2D algorithm with a rectangular grid (as described in Section~\ref{sec:2d}) whose aspect ratio mirrors the minor differences between $d$ and $f$. In particular, this
%flexible 2D algorithm asymptotically communicates no more data volume than any of the 1.5D algorithms described in the earlier work~\cite{spdmmm16}. 

%\input{partscheme2}
%\input{partscheme3}

\subsection{Block Two-Dimensional (2D) Algorithms}
\label{sec:2d}
\hide{
\begin{table}[!h]
\centering
\caption{Data distribution for block 2D algorithms}
\begin{tabular}{ |p{2cm}||p{3cm}|}
\hline
Matrix & Partitioning  \\
\hline
$\mA$ & Block 2D \\
$\mH^l$ & Block 2D \\
$\mG^l$ & Block 2D \\
$\mW^l$ & Fully-replicated \\
\hline
\end{tabular}
\end{table}}

Processes are logically organized on a square 
$P = P_r \times P_c$ mesh, indexed by their row and column indices so that the $(i,j)$th process is denoted by $P(i,j)$. 
Matrices are assigned to process according to a 2D block decomposition. For a given $n \times m$ matrix, each process gets a submatrix of dimensions 
$n / P_r \times m / P_c$ in its local memory. For example, $\mA\transpose$ is partitioned as shown below and $\mA\transpose_{ij}$ is assigned 
to process $P(i,j)$. 

\begin{equation}
\mA\transpose = \left( 
\begin{array}{c c c}
\mA\transpose_{11} & \ldots  & \mA\transpose_{1 p_c} \\
%\hline
\vdots  & \ddots & \vdots  \\
%\hline
\mA\transpose_{p_r 1} & \ldots   & \mA\transpose_{p_r p_c} \\ 
\end{array} 
\right)\label{eqn:2dpartitioning}
\end{equation}

The pseudocode for the general rectangular case is listed in Algorithm~\ref{alg:sparsesumma}. \revision{The first phase of the algorithm is based on a variant of the Scalable Universal Matrix Multiplication (SUMMA) algorithm~\cite{van1997summa}.} When \proc{BCast}$(\mA_{ic}, P(i,:))$ is called, the owner of $\mA_{ic}$ becomes the root and broadcasts
its submatrix to all the processes along the $i$th process row. Symmetrically, \proc{BCast}($\mH_{rj}, P(:,j)$) means that whomever is the owner of $\mH_{rj}$ broadcasts its submatrix along the $j$th process column.

Variables $\mathit{lcols}$ and $\mathit{lrows}$, which are significant only at the broadcasting processes, are the local column and row ranges for matrices that are to be broadcast. Here we used the colon notation where the range $i,i+1,\ldots,i+k$ is denoted by $(i:i+k)$ and empty colon $(:)$ specifies the full possible range. For example, $M(:,j)$ denotes the $j$th column, and $M(1:k,:)$ denotes the first $k$ rows of any two-dimensional object $M$. The \proc{SpMM} and \proc{GEMM} are local sparse-dense and dense-dense matrix multiplications, respectively. 

\begin{algorithm}
\begin{algorithmic}[1]
%\Require $\mA: \mathbb{R}^{n \times n}:$ sparse adjacency matrix, 
%$\mH^l: \mathbb{R}^{n \times f^{l-1}}:$ dense input activations matrix, 
%$\mW: \mathbb{R}^{f^{l-1} \times f^{l}}:$ dense training matrix
%\Ensure $\mH^l: \mathbb{R}^{n \times f^l}:$ dense output activations matrix.
\Procedure{Block2DFW}{$\mA\transpose, \mH^{l-1}, \mW, \mH^l$}
\For{all processes $P(i,j)$ \InParallel} 
	\For{ $q =1$ to $f^{l-1}/b$} \Comment{$1$st SUMMA phase}
	\State $c = (q b)/p_c$ \Comment{broadcasting process column}
	\State $r = (q b)/p_r$ \Comment{broadcasting process row}
	\State $\id{actv} = (q b : (q+1) b)$ \Comment{active columns/rows}
 	\State $\id{lcols} = \id{actv} \pmod{(n/p_c)}$  
	\State $\id{lrows} = \id{actv} \pmod{(n/p_r)}$ 
	\State	$\hat{\mA}\transpose \gets $ \Call{BCast}{$\mA\transpose_{ic}(:,\id{lcols}), P(i,:)$} 
	\State	$\hat{\mH}^{l-1} \gets $ \Call{BCast}{$\mH^{l-1}_{rj}(\id{lrows},:), P(:,j)$} 		
	\State	$\mT_{ij} \gets \mT_{ij} + \Call{SpMM}{\hat{\mA}\transpose,\hat{\mH}^{l-1}}$	
	\EndFor
	\For{ $q =1$ to $f^{l-1}/b$} \Comment{$2$nd phase}
	\State $c = (q  b)/p_c$ 
	\State $\id{actv} = (q b : (q+1) b)$
    \State $\id{lcols} = \id{actv} \pmod{(n/p_c)}$ 
    \State	$\hat{\mT} \gets $ \Call{BCast}{$\mT_{ic}(:,\id{lcols}), P(i,:)$} 
    \State $\id{colrange} = (j f^l/p_c +1: (j+1) f^l/p_c)$ 
	\State	$\mH^l_{ij} \gets \mH^l_{ij} + \Call{GEMM}{\hat{\mT},\mW(\id{actv}, \id{colrange})}$	
	\EndFor
\EndFor
\EndProcedure
\end{algorithmic}
\caption{Block 2D algorithm for GNN forward propagation, which computes $\mH^{l} \gets \sigma(\mA\transpose \mH^{l - 1}\mW^l)$ in parallel. $\mA$ and $\mH$ are distributed on a $p_r \times p_c$ process grid, $\mW$ is replicated. Blocking parameter $b$ is required to evenly divide $f^{l-1}/p_r$ and $f^{l-1}/p_c$. } \label{alg:sparsesumma}
\end{algorithm}

We start by analyzing the special $p_r = p_c = \sqrt{p}$ case to give the intuition. For each process $P(i,j)$, the computation of the intermediate product $\mT = \mA\transpose \mH^{l - 1}$ is:
 $$ \mT_{ij} = \sum _{k=1}^{\sqrt p} \mA\transpose_{ik} \, \mH_{kj}$$

\subsubsection{Equation  $\mZ^l = \mA\transpose \mH^{l - 1}\mW^l$}
% \begin{figure}[h]
%     \centering
%     \includegraphics[scale=0.4]{ps4eq1.png}
%     \caption{Partitioning Scheme $4$ Equation $1$}
%     \label{fig:ps3eq1}
% \end{figure}
\paragraph*{Communication: 2D SUMMA SpMM + partial SUMMA} 
    Both $\mA^T$ and $\mH^{l - 1}$ are partitioned into a $\sqrt{P} \times \sqrt{P}$ process grid. This yields a $2$D multiplication which we can do with an optimized SUMMA algorithm. To compute a submatrix of $\mA^T\mH^{l - 1}$, each process in the submatrix's row must broadcast their $\mA$ and each submatrix's column must broadcast their $\mH^{l - 1}$. $\mW^l$ is fully-replicated on each process, and is multiplied with $\mT = \mA^T\mH^{l - 1}$ after communication. However, this also requires communicating $n\times f$ sized $\mT$ along the process row, something we label as ``partial SUMMA". 
    \begin{align*}
        T_{comm} = \alpha 2 \sqrt{P} + \beta \Big(\frac{\dnnz(\mA)}{\sqrt{P}} + \frac{2nf}{\sqrt{P}}\Big) 
    \end{align*}

\subsubsection{Equation $\mH^l = \sigma(\mZ^l)$}
% \begin{figure}[h]
%     \centering
%     \includegraphics[scale=0.4]{ps4eq2.png}
%     \caption{Partitioning Scheme $4$ Equation $2$}
%     \label{fig:ps3eq1}
% \end{figure}

\paragraph*{Communication: All-Gather} 
    $\mH^l$ is partitioned in a 2D process grid. When $\sigma$ is element-wise, no communication is needed. However, when $\sigma$ is not element-wise, in particular for \texttt{log\_softmax}, each process needs to broadcast its $\mH^{l - 1}$ with the entire process row.
    \begin{align*}
        T_{comm} =  \alpha \lg{P} + \beta \frac{nf}{\sqrt{P}} 
    \end{align*}
\subsubsection{Equation $\mG^{l-1} = \mA \mG^{l}(\mW^{l})\transpose \odot \sigma '(\mZ^{l-1})$}

%\begin{figure}[h]
%    \centering
%    \includegraphics[scale=0.4]{ps4eq3.png}
%    \caption{Partitioning Scheme $4$ Equation $2$}
%    \label{fig:ps3eq2}
%\end{figure}
\paragraph*{Communication: 2D SUMMA SpMM + partial SUMMA + All-Gather} 
    $\mA$ and $\mG^{l}$ are partitioned in a 2D process grid, and results in the same communication pattern as Equation 1. We also need to account for $\sigma' (\mZ^{l-1})$. Recall that this needs communication when $\sigma$ is not elementwise. This has the same communication pattern as Equation 2. The communication cost is 
    \begin{equation*}
         T_{comm} = \alpha (2\sqrt{P} +\lg{P}) + \beta \Big(\frac{\dnnz(\mA)}{\sqrt{P}} + \frac{3nf}{\sqrt{P}}\Big).
    \end{equation*}

\subsubsection{Equation $\mY= (\mH^{l - 1})\transpose \mA \mG^l$}
% \begin{figure}[h]
%     \centering
%     \includegraphics[scale=0.4]{ps4eq4.png}
%     \caption{Partitioning Scheme $4$ Equation $4$}
%     \label{fig:ps2eq3}
% \end{figure}
\paragraph*{Communication: 2D dense SUMMA + All-Gather} 
    Strictly speaking, there are two matrix multiplications here. The first is multiplying $\mA$ and $\mG^l$, which would have been a 2D SUMMA SpMM. \revision{However, we can reuse the same intermediate product from the previous equation, as we have done in the case of 1D Block Row algorithm. This increases storage by an additive $n f / P$ term on each process. The intermediate product $\mA \mG^l$ is already partitioned on a $\sqrt{P} \times \sqrt{P}$ process grid. The second multiplication, which is the only multiplication we have to pay for computing this equation, is between $(\mH^{l - 1})\transpose$ and the previously saved intermediate product $\mA \mG^l$.} This is a 2D dense SUMMA on two matrices with $nf$ elements, resulting in a small $f\times f$ output. The final allgather is to keep $\mY$ replicated. Overall, the communication cost due to this equation is
    \begin{equation*}
         T_{comm} = \alpha (\sqrt{P} + \lg{P}) + \beta \Big(\frac{2nf}{\sqrt{P}} + f^2\Big).
    \end{equation*}

\subsubsection{Total Communication}

\label{sec:total2D}
\begin{align*}
    &= \sum_{l = 1}^{L}\Big(\alpha(5\sqrt{P} + 3\lg{P}) + \beta\big(\frac{8nf^l}{\sqrt{P}} + \frac{2\dnnz(\mA)}{\sqrt{P}}+(f^l)^2\big)\Big) \\
            &\approx L \Big(\alpha(5\sqrt{P} + 3\lg{P}) + \beta\big(\frac{8nf}{\sqrt{P}} + \frac{2\dnnz(\mA)}{\sqrt{P}} + f^2\big)\Big)
\end{align*}
The communication volume scales with $\sqrt{p}$. However, the constants in the 2D algorithm are significantly larger than the constants in 1D and 1.5D algorithms. The latency cost of the 2D algorithm is asymptotically better than the 1D algorithm but worse than the 1.5D algorithm when $c{=}\sqrt{p}$. Finally, the 2D algorithm also moves the sparse matrix, creating additional costs. This is not a problem asymptotically as long as $\dnnz(A) = O(nf)$ but can be a bottleneck if the graph is larger than the aggregate embedding size across all vertices.  

\subsubsection{The Rectangular Grid Case}
When the process grid is non-square, the 2D SUMMA algorithm is still well-defined and relatively easy to implement. Consider the forward propagation equation where $\mA\transpose$ and $\mH$ are 2D block partitioned on a $P_r \times P_c$ grid. Each process needs to access $(1/P_r)$th of $\mA\transpose$ and $(1/P_c)$th of $\mH$ to form its own $n/P_r \times f/P_c$ piece of the intermediate output $\mT = \mA\transpose \mH$. However, we now need to consider the communication due to the 2nd phase of Algorithm~\ref{alg:sparsesumma}, which communicates this $\mT$ matrix along the process row. The forward propagation communication cost becomes:
\begin{align*}
        T_{comm} = \alpha \, \text{gcf}(P_r, P_c) + \beta \Big(\frac{\dnnz(\mA)}{P_r}+ \frac{nf}{P_c}+\frac{nf}{P_r}\Big),
\end{align*}
    
\noindent
where gcf denotes greatest common factor. This suggests that if the average vertex degree of the graph is significantly larger than the feature vector length, then there are potential savings is terms of sparse matrix communication by increasing the $P_r/P_c$ ratio. However, this comes at the expense of increasing the sum of other two terms, which correspond to dense matrix communication. This is because the sum of two numbers whose product is fixed is minimized when those numbers are equal, or put differently, square has the smallest perimeter of all rectangles of a given area. Consequently, given the unclear benefit to cost ratio of using non-square grids, our implementations focus on square grids in this work. 

\subsection{Block 3D algorithms}
\label{sec:3d}

\hide{
\begin{table}[!h]
\caption{Data distribution for block 3D algorithms}
\centering
\begin{tabular}{ |p{2cm}||p{3cm}|}
\hline
Matrix & Partitioning  \\
\hline
$\mA$ & Block Split 3D \\
$\mH^l$ & Block Split 3D \\
$\mG^l$ & Block Split 3D \\
$\mW^l$ & Fully-replicated \\
\hline
\end{tabular}
\end{table}}

For ease of presentation, let us assume that processes are logically organized on a 3D 
$\sqrt[3]{P} \times \sqrt[3]{P} \times \sqrt[3]{P}$ mesh, indexed by three indices, though in general each of the three process dimensions can be different.
Our matrix to process mesh assignment follows the Split-3D-SpGEMM approach of Azad et al.~\cite{Azad2016}. We call our variation Split-3D-SpMM, because the primary GNN operation is SpMM as opposed to SpGEMM. Each 2D plane in this 3D process mesh is called a ``layer''. 

Considering a single SpMM such as the $\mA \mH^{l-1}$ in forward propagation, we note that two input matrices are split  differently. 
After 3D distribution, each local submatrix $\mA_{ijk}$ of $\mA$ is of dimensions $n/\sqrt[3]{P}\times n/\sqrt[3]{P}^2$.
That means the number of rows of each $\mA_{ijk}$ is $\sqrt[3]{P}$ times its number of columns. By contrast, $\mH$ is split along the rows across layers and each local piece $\mH_{ijk}$ is of dimensions $n/\sqrt[3]{P}^2\times f/\sqrt[3]{P}$. This data distribution choice makes each $\mH^{l-1}_{ijk}$ shorter and fatter than 2D distribution makes, potentially alleviating some of the issues with local SpMM scaling we observe with our 2D implementation. 

\revision{We note that each process only holds $(1/P)$th of the input matrices in Split-3D-SpMM, so there is no replication at the input stage. The replication happens in intermediate stages, as we explain in detail below.}

\subsubsection{Equation $\mZ^l = \mA\transpose \mH^{l - 1}\mW^l$}
% \begin{figure}[h]
%     \centering
%     \includegraphics[scale=0.4]{ps4eq1.png}
%     \caption{Partitioning Scheme $4$ Equation $1$}
%     \label{fig:ps3eq1}
% \end{figure}
\paragraph*{Communication: One full and one partial Split-3D-SpMM}
    The easiest way to think about a 3D multiplication algorithm is to consider it as independent 2D algorithms executing at each layer, followed by a reduction. 
    To compute a submatrix of $\mA\transpose\mH^{l - 1}$, each submatrix in $\mA\transpose_{:jk}$ broadcasts itself to the rest of the process row, and each submatrix $\mH_{i:k}$ broadcasts itself to the rest of the process column.
    In each 2D SUMMA iteration, each process on a given layer receives a submatrix from $\mA^T$ and a submatrix from $\mH^{l - 1}$, multiplies them, and adds them to a running sum. 
    
    Once these 2D SUMMA iterations that have been executing independently at each layer complete, processes have partial sums for $\mA^T\mH^{l - 1}$ that need to be reduced across the third process dimension (also called a ``fiber''). \revision{The partial sums after the 2D SUMMA iterations complete are $n/\sqrt[3]{P} \times  f/\sqrt[3]{P}$ dense matrices, with $nf/P^{2/3}$ elements each. These partial sums are then reduce-scattered along the fiber dimension to get the product $\mA\transpose\mH^{l - 1}$ in a Block Split 3D format.} This results in the following communication cost just to form $\mA\transpose\mH^{l - 1}$:
    \begin{align*}
        T_{comm} = \alpha (P^{1/3}+ \lg{P}) + \beta \Big(\frac{\dnnz(\mA)}{P^{2/3}} +  \frac{2nf}{P^{{2/3}}}\Big)
    \end{align*}

    \revision{Note that the aggregate memory consumption over all processes
    prior to the fiber reduction would be $P \big( nf /P^{2/3} \big) =  P^{1/3} nf$, where $P^{1/3}$ is the well-known memory replication cost factor of 3D algorithms~\cite{agarwal1995three, ballard2011minimizing}.} 
    
    We now need to multiply this intermediate product $\mA\transpose\mH^{l - 1}$ with $\mW^l$. Similar to the Block 2D case, we will perform a partial 3D dense matrix multiply where the second input matrix does not need to be communicated because it is replicated. The total communication cost, with this partial step added, is:
    \begin{align*}
        T_{comm} = 2 \alpha (P^{1/3}+ \lg{P}) + \beta \Big(\frac{\dnnz(\mA)}{P^{2/3}} +  \frac{4nf}{P^{{2/3}}}\Big)
    \end{align*}

\subsubsection{Equation $\mH^l = \sigma(\mZ^l)$}
% \begin{figure}[h]
%     \centering
%     \includegraphics[scale=0.4]{ps4eq2.png}
%     \caption{Partitioning Scheme $4$ Equation $2$}
%     \label{fig:ps3eq1}
% \end{figure}

\paragraph*{Communication: All-Gather} 
    $\mH^l$ is partitioned in a 3D process mesh. When $\sigma$ is elementwise, no communication is needed. However, when $\sigma$ is not elementwise, in particular for \texttt{log\_softmax}, each process needs to broadcast its $\mH^l$ with the entire process row within a layer. This is equivalent to an all-gather per layer. No cross-layer or cross-row communication is needed as the output of \texttt{log\_softmax} for a row of $\mZ$ is only dependent on the values within that row.
    \begin{align*}
        T_{comm} = \alpha\lg{P} + \beta \Big(\frac{nf}{P^{{2/3}}}\Big)
    \end{align*}

\subsubsection{Equation $\mG^{l-1} = \mA \mG^{l}(\mW^{l})\transpose \odot \sigma '(\mZ^{l-1})$}
%\begin{figure}[h]
%    \centering
%    \includegraphics[scale=0.4]{ps4eq3.png}
%    \caption{Partitioning Scheme $4$ Equation $2$}
%    \label{fig:ps3eq2}
%\end{figure}
\paragraph{Communication: One full and one partial Split-3D-SpMM, and All-Gather} 
    $\mA$ and $\mG^{l}$ are partitioned in a 3D process grid, and results in the same communication pattern as Equation 1. We also need to account for $\sigma' (\mZ^l)$. Recall that this needs communication when $\sigma$ is not elementwise. This has the same communication pattern as Equation 2, hence totalling:
    
    \begin{align*}
        T_{comm} = \alpha (2 P^{1/3} + 3 \lg{P}) + \beta \Big(\frac{\dnnz(\mA)}{P^{2/3}} +  \frac{5nf}{P^{{2/3}}}\Big)
    \end{align*}

\subsubsection{Equation $\mY= (\mH^{l - 1})\transpose \mA \mG^l$}
% \begin{figure}[h]
%     \centering
%     \includegraphics[scale=0.4]{ps4eq4.png}
%     \caption{Partitioning Scheme $4$ Equation $4$}
%     \label{fig:ps2eq3}
% \end{figure}
\paragraph*{Communication: 3D dense SUMMA + All-Gather} 
    \revision{We store the intermediate product $\mA \mG^l$ that has been computed in the previous step, and reuse it here as we have done in other algorithms.} Multiplying $(\mH^{l - 1})\transpose$ with $\mA \mG^l$ is a dense 3D SUMMA on two matrices with $nf$ elements, resulting in a small $f \times f$ output. Note that each of these inputs are Block Split 3D. As before, the final allgather is to keep $\mY$ replicated. The bandwidth cost of all-gather strictly dominates the bandwidth cost of fiber reduction, so we do not include the cost of fiber reduction in the bandwidth term. The resulting communication cost is: 
        \begin{align*}
    T_{comm} = \alpha 2\lg{P} + \beta \Big(\frac{2nf}{P^{2/3}} + f^2\Big).
    \end{align*}

\subsubsection{Total Communication} Ignoring $\lg{P}$ latency terms that are strictly dominated by the $P^{1/3}$ terms, we have
\begin{align*}
    T_{comm} \approx L \Big(\alpha(4 P^{1/3}) + \beta\Big(\frac{2 \dnnz(\mA)}{P^{2/3}} + \frac{12nf}{P^{2/3}}\Big)\Big)
\end{align*}

Although the 3D algorithm provides an asymptotic reduction in communication costs, it has several disadvantages compared to the 2D algorithm, which are (1) its high constants, (2) its implementation complexity, and (3) its need to do a factor of $\sqrt[3]{p}$ replication in its intermediate stages. 

\revision{The exact impact of this intermediate replication to the overall memory consumption of GNN training depends on several factors such as (1) the number of layers, (2) the ratio of the maximum number of activations in a layer to the total number of activations, and (3) the ratio of the number of total activations to the number of edges (i.e. $\dnnz(\mA)$) in the graph. An implementation of full-batch GNN training that is optimal with respect to memory usage has the following memory footprint:}
\revision{
\begin{align*}
    M^*_{total} = \dnnz(\mA) + \sum_{l = 1}^{L}\big( n f^l \big) \approx  \dnnz(\mA) + nf L
\end{align*}
}
\revision{
The memory consumption of the 2D algorithm asymptotically matches this bound and the only overheads are in the communication buffers. However, the 3D algorithm replicates the activations while processing a layer. The important point is that this replication only impacts the current layer that is being processed, either during forward or backward propagation, because the intermediate matrices are discarded after layer-wise reduction. Suppose the maximum number of activations (e.g., features) is $f^{max} = \max_{l=1}^L f^l$, and the layer with the highest number of activations is $k = \argmax_{l=1}^L f^l$. Then the memory footprint of the 3D algorithm is}
\revision{
\begin{align*}
    M^{3D}_{total} &= \dnnz(\mA) + \sum_{l = 1, l \neq k}^{L}\big( n f^l \big) + \sqrt[3]{p} \, nf^{max} \\
                &\approx  \dnnz(\mA) + nf (L-1) + \sqrt[3]{p} \, nf^{max} 
\end{align*}}
\revision{
When the network is deep or when the maximum number of activations in a layer is much smaller than the input graph size, this memory overhead is unlikely to be an impediment. However, many existing GNN networks are rather shallow at the moment. }

\section{Experimental Setup}
\label{sec:exp-set}
\subsection{Datasets and Compute Platform}
\label{sec:data}
We ran our experiments on two of the largest datasets used in GNN research previously, the Reddit and Amazon datasets. In addition, we use a much larger protein similarity network which pushes the boundaries of large-scale GNN traning. This `protein' network comes from the data repository of the HipMCL algorithm~\cite{hipmcl}. It is an induced subgraph, which contains 1/8th of the original vertices, of the sequence similarity network that contained all the isolate genomes from the IMG database at the time. The characteristics of the datasets are documented in Table~\ref{tab:data}. The feature (input) and label (output) embedding lengths of the protein dataset largely follows the literature on this topic~\cite{gligorijevic2019structure} where the protein sequences are assumed to be initially processed independently to obtain their 128-length embeddings (via CNNs or LSTMs) before running GNNs on those embeddings. \revisiontwo{For load balancing, graph vertices are randomly permuted prior to training.}

\begin{table}[!h]
\centering
\caption{Datasets used in our experiments}
\label{tab:data}
\begin{tabular}{ |l|r|r|r|r|}
\hline
Name & Vertices & Edges & Features & Labels  \\
\hline
Reddit & 232,965 & 114,848,857 & 602 & 41\\
Amazon & 14,249,639 & 230,788,269 & 300 & 24 \\
% Protein & 8,745,542 & 1,309,240,502 & 128 & 256 \\
Protein & 8,745,542 & 2,116,240,124 & 128 & 256 \\
\hline
\end{tabular}
\end{table}

We use the same 3-layer GNN architecture presented by Kipf and Welling~\cite{KipfWelling2017} though deeper and wider networks are certainly possible as done by ROC~\cite{mlsys2020_83} given the similar performance we achieve.

We verified that our parallel implementation not only achieves the same training accuracy in the same number of epochs as the serial implementations in PyTorch, but it also outputs the same embeddings up to floating point accumulation errors. Consequently, we only provide performance numbers as the accuracy numbers are identical to the serial cases.

\subsection{System details}
\label{sec:system}
All of our experiments are run on the Summit supercomputer at ORNL, which has IBM AC922 nodes with 6 NVIDIA V100 GPUs. Each GPU has 16GB of HBM2 memory. Each Summit node has two sockets, each with 1 POWER9 CPU and 3 GPUs each. Within a socket, each GPU is connected to each other and the host CPU with NVLINK 2.0 with $50$\,GB/s unidirectional bandwidth. Across sockets, Summit uses the IBM X-bus interconnect with $64$\,GB/s. Each socket has a $16$\,GB/s link to the Network Interface Card (NIC). Across nodes, Summit uses dual-rail EDR Infiniband with $25$\,GB/s node injection bandwidth~\cite{summittalk}, but each socket has access to half ($12.5$\,GB/s) of this node injection bandwidth. 

\subsection{Implementation details}
% \Alok{talk about PyTorch geometric, Pytorch distributed, NVCC, how you implemented the code - directly in python mostly, except for calling cuSPARSE's csrmm2, etc,etc }
We implement our 3-layer GNN architecture mostly in PyTorch Geometric (PyG) 1.3 \cite{Fey/Lenssen/2019}. Within PyG, we use torch.distributed with a NCCL backend for our communication primitives. \revisiontwo{NCCL implements broadcasts and reductions using a ring algorithm, and splits an outgoing message to smaller chunks for pipelining~\cite{ncclrepo}. NCCL collectives have different completion semantics than MPI. For example, when the broadcasting process receives the delivery confirmation of its own data from its peer process, its function call returns without waiting for the completion of the broadcast on subsequent nodes. This relaxed completion combined with pipelining allows broadcasting using NCCL to be as high-performance as using manual shifting when implementing matrix multiplication, provided that the messages can be divided into enough chunks to hide the latency of filling the pipeline.} 

For our SpMM calls, we separately call cuSPARSE's \texttt{csrmm2} function in a C++ extension. We compile our C++ backend with CUDA 10.1. Recall each node on Summit has 6 GPUs. As such, our implementation is single-node multi-GPU when $P{=}4$, but multi-node multi-GPU in all other cases. \revisiontwo{In particular, a $P$ process job allocates $\lceil P/6 \rceil$ nodes so that we utilize all 6 GPUs except for the last node. We only deviate from this setup for exposing the topological limitations of Summit interconnect in Figure~\ref{fig:15dresults1GPU}} 

\begin{figure*}[!t]
    \centering
    \includegraphics[width=\textwidth,height=5cm]{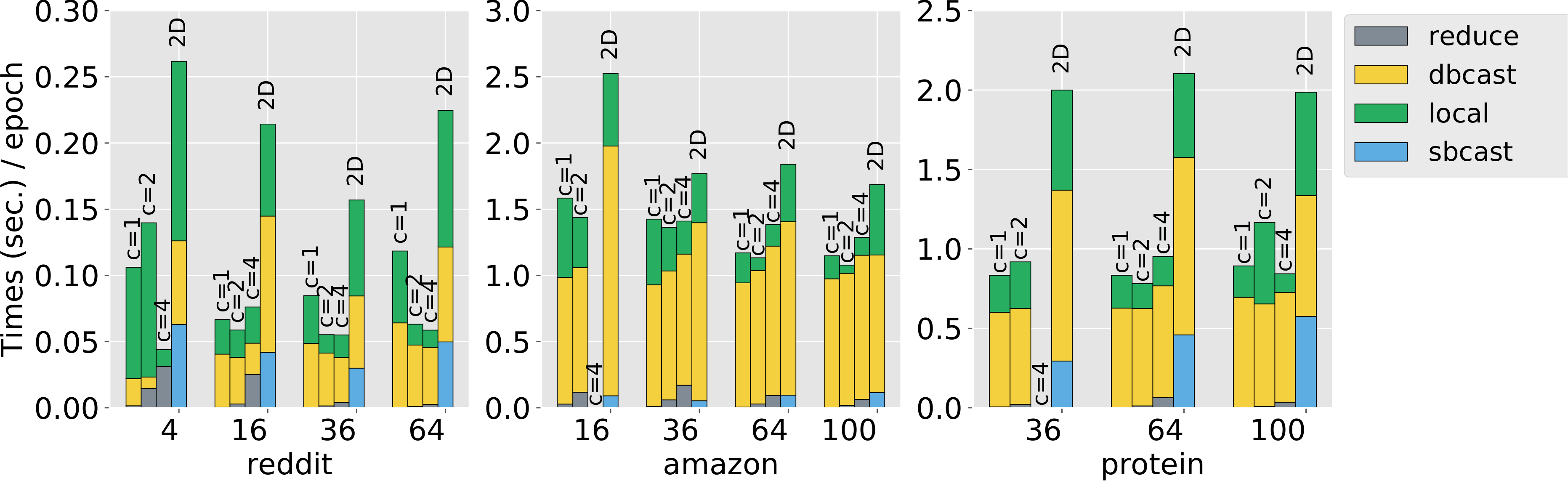}
    \caption{1D ($c{=}1$), 1.5D ($c{=}2,4$), and 2D performance results when using all 6 GPUs on each node. The x-axis in each subplot is the number of GPUs used. \textit{dbcast} refers to the broadcast of dense embedding matrices, \textit{sbcast} refers to the broadcast of sparse adjacency matrix (only for 2D), \textit{reduce} is the allreduce (only for 1.5D), \textit{local} is the local computation including cuSPARSE SpMM calls, small DGEMM calls, transpose (only for 2D), and sparse matrix assembly after communication (only for 2D). Missing bars for $c{=}4, p{=}16$ on Amazon and $c{=}4, p{=}36$ on Protein means that those runs ran out of memory.}
    \label{fig:perf-breakdown}
\end{figure*}

For Reddit, we use the input feature vectors and training split used by Hamilton et~al.~\cite{hamiltonInductive2017} as they are already provided within PyG. 
For the Amazon and Protein datasets, we randomly generate feature values for simplicity and use the whole graph as the training set. This does not affect performance, and in practice, users could use any values for the feature vectors. We run Reddit and Amazon for $100$ epochs and Protein for $10$ epochs. We do not report numbers for Amazon on $4$ devices or numbers for Protein on $4$ or $16$ devices as the data does not fit in memory for those configurations. Jia et~al.~\cite{mlsys2020_83} observed the same behavior with PyG. We are unable to compare our results directly with Neugraph or ROC because neither code is currently publicly available.

\hide{
\section{Results}
\begin{figure}[!t]
    \centering
    \includegraphics[scale=0.28]{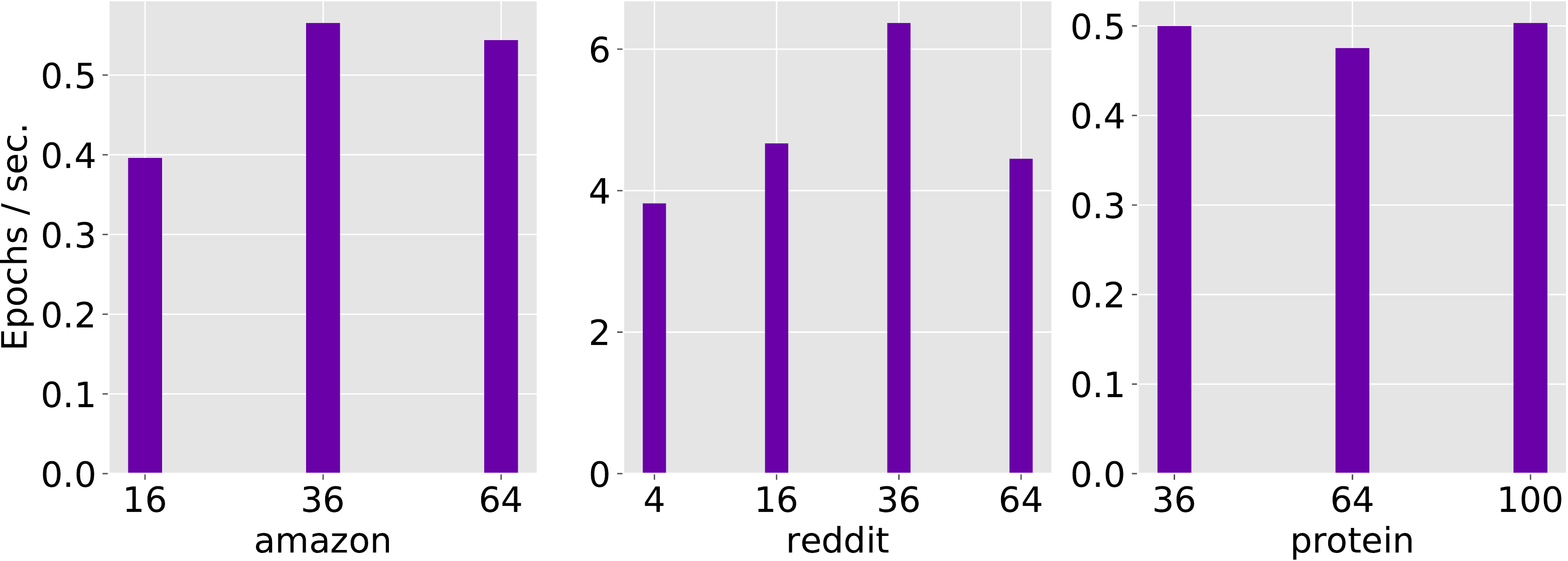}
    \caption{Epoch Throughput of 2D Implementation across GPU Counts for a Single Process \Aydin{Is this plot still needed? It doesn't add new information as readers can invert Fig 3 in their heads; it just takes space.} \Alok{Yeah probably fine to get rid of it}}
    \label{fig:epoch-through}
\end{figure}}

\section{Results}
\subsection{Performance of the 1D and 1.5D Implementations}
\hide{
\begin{figure}[!t]
    \centering
    \includegraphics[scale=0.28]{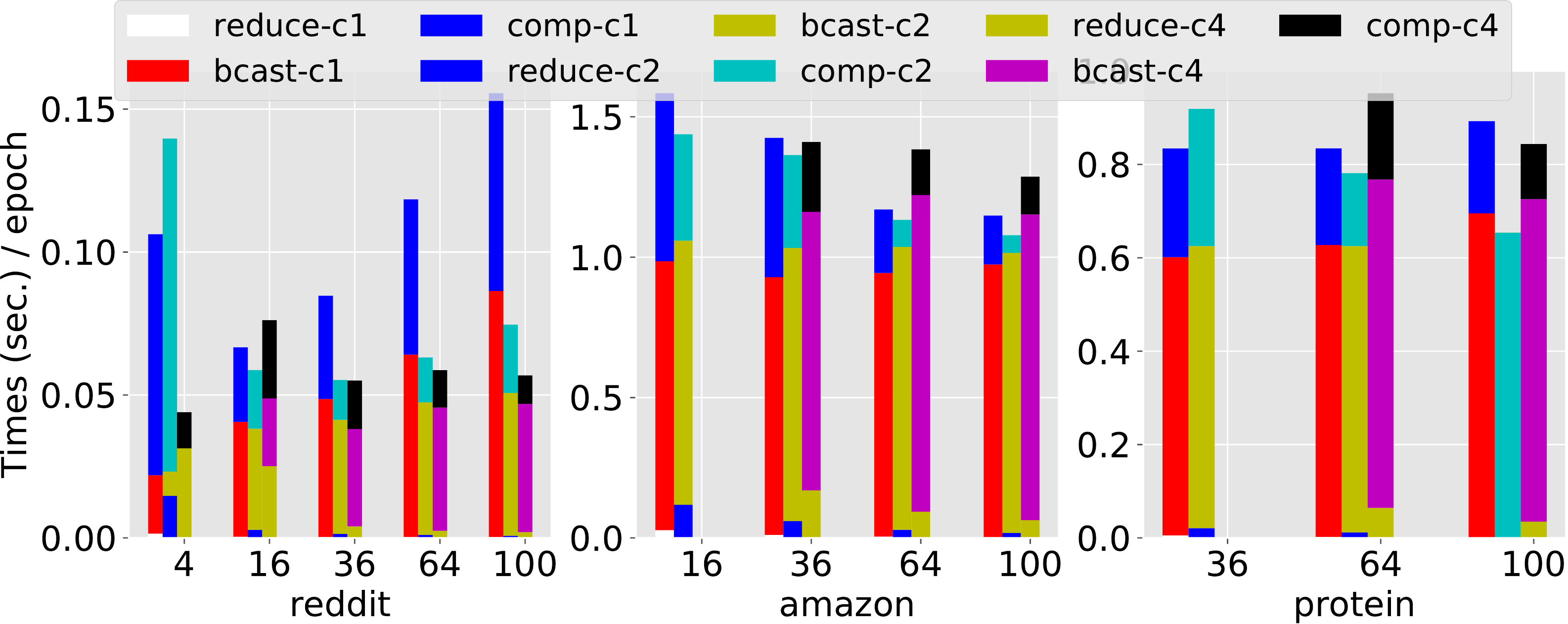}
    \caption{1.5D Implementation results}
    \label{fig:15dresults}
\end{figure}
}

\begin{figure}[!t]
    \centering
    \includegraphics[scale=0.28]{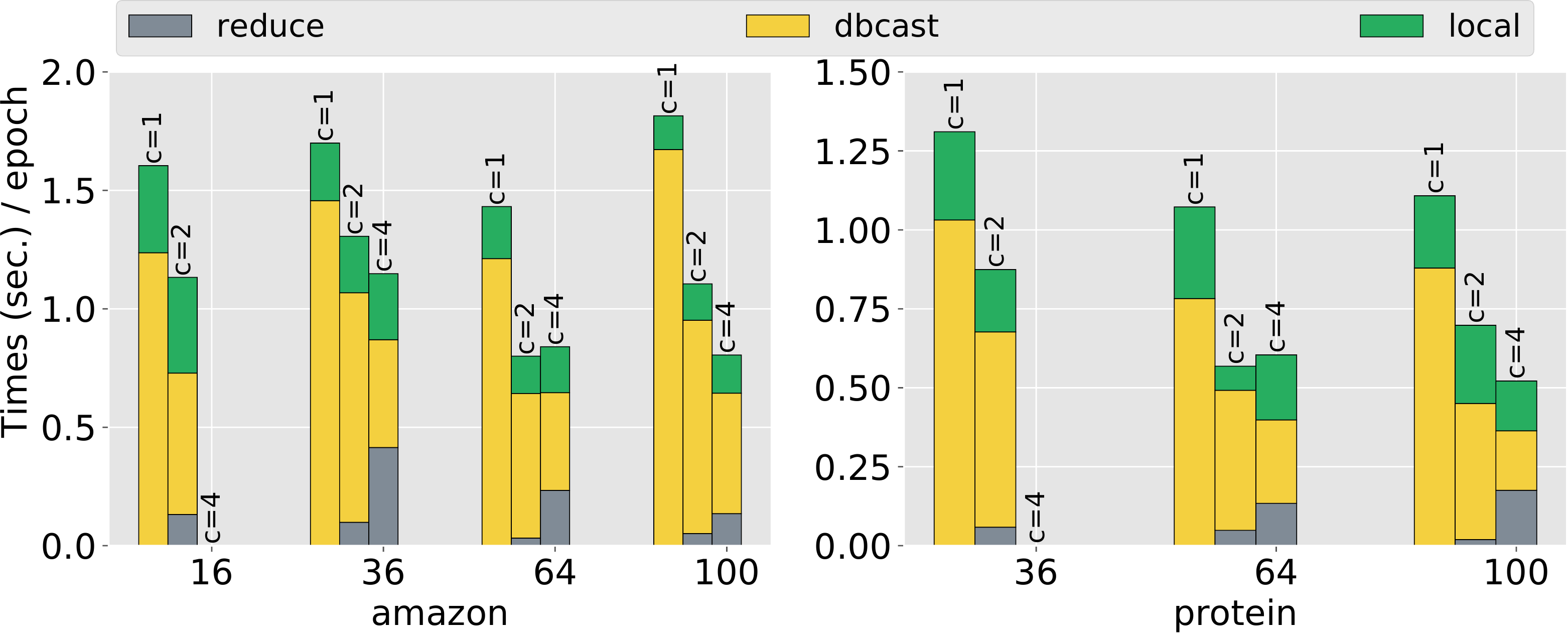}
    \caption{1D ($c{=}1$), 1.5D ($c{=}2,4$) performance results when only one 1 GPU is used per node. The x-axis for each subplot is the number of GPUs used. \textit{dbcast} refers to the broadcast of dense embedding matrices, \textit{reduce} is the allreduce (only for 1.5D), \textit{local} is the local computation including cuSPARSE SpMM calls, small DGEMM calls}
    \label{fig:15dresults1GPU}
\end{figure}
\revisiontwo{The performance of 1D ($c{=}1$) as well as 1.5D implementations ($c{>}1$) are shown in Figure~\ref{fig:perf-breakdown}. Since our 1D and 1.5D implementations only move the dense matrices, the communication volume is proportional to the product of the number of vertices and the number of features. Due to its small vertex count, GNN training on Reddit dataset is increasingly latency bound at large concurrencies. Consequently when $P$ is large, increasing $c$ directly translates into lower communication costs for Reddit, due to quadratic decrease in latency costs. On a single node, our 1.5D GNN training algorithm achieves more than $20$ epochs/sec throughput, higher than all previously published results. }

\revisiontwo{For Amazon and Protein, which are mostly bandwidth bound, our analysis expects communication volume in the broadcast stage to decrease linearly with increasing $c$. However, our results showed minimal decrease in broadcast time when fully utilizing all 6 GPUs on each Summit node. Diving into the specifics of Summit architecture~\cite{summittalk}, as explained in Section~\ref{sec:system}, we conclude this is due to sharing network injection bandwidth. Recall that each socket on Summit has 3 GPUs and they share the same $12.5$\,GB/s  network injection bandwidth. Also recall that broadcasts in NCCL are implemented using a pipelined ring algorithm. When only a single broadcast is active ($c{=}1$ aka 1D), the whole $12.5$\,GB/s injection bandwidth is used by a single GPU because the last GPU on the $i$th node is peered with the first GPU on the $(i+1)$th node. When $c=2$, there are two simultaneous broadcasts happening on two virtual rings. Two GPUs on a socket now has to share the network injection bandwidth to communicate with their peers on the neighboring node. This cuts down the effective available bandwidth by half, so even though the communication volume is reduced to half we do not see any appreciable decrease in broadcast times with increasing $c$ from $1$ to $2$.}

\revisiontwo{ Further increasing $c$ to $4$ increases the load node injection to its maximum where all 3 GPUs on the socket compete for $12.5$\,GB/s injection bandwidth. We confirmed that node injection bandwidth is indeed the bottleneck by running our 1D and 1.5D implementations on 1 GPU/node configuration. Figure~\ref{fig:15dresults1GPU} shows close-to-linear scaling for broadcast times when $c$ is increased, for the bandwidth-bound datasets Amazon and Protein.}

\hide{
\begin{figure}[!t]
    \centering
    \includegraphics[scale=0.28]{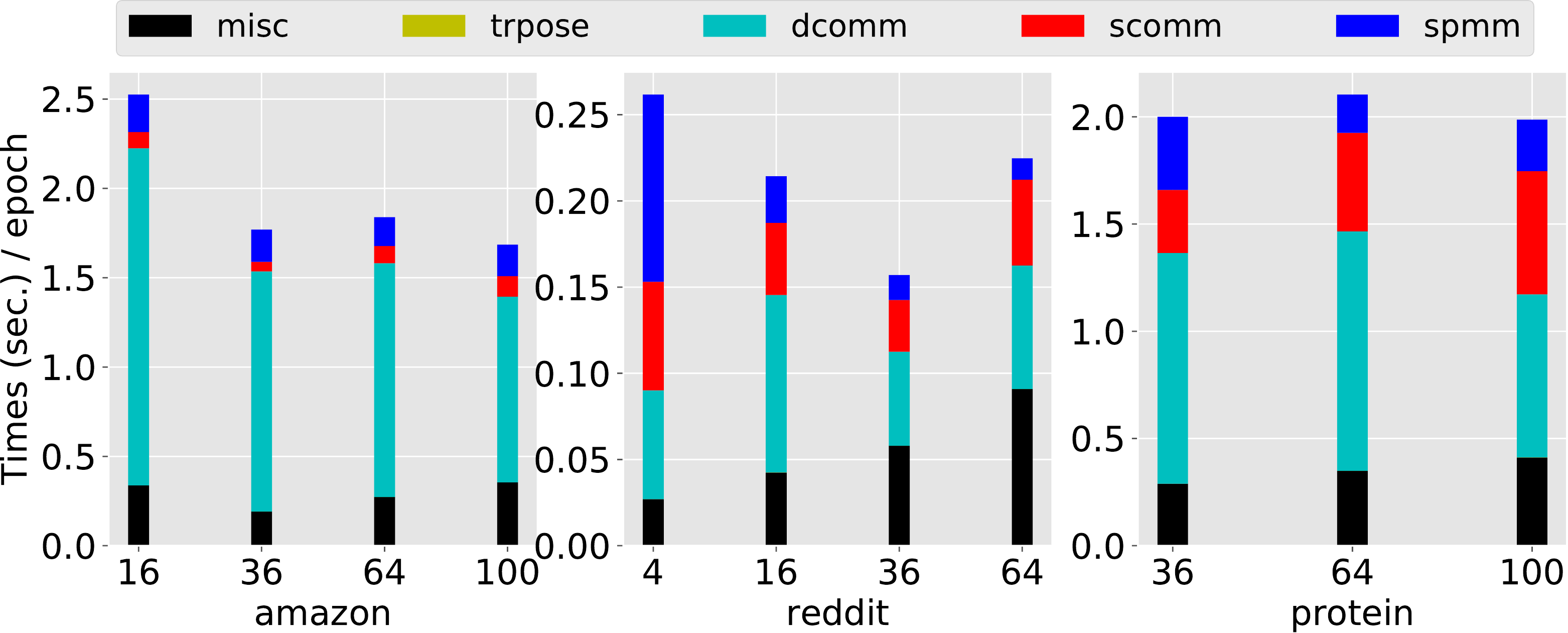}
    \caption{Performance breakdown of 2D implementation across GPU counts for a single process. \textit{scomm} refers to communicating sparse matrices, \textit{dcomm} refers to communicating dense matrices, and \textit{trpose} refers to computing matrix transposes. \textit{spmm} refers to local spmm calls that involve no communication. Local dense matrix multiply (\texttt{GEMM}) calls are inexpensive and thus reported under \textit{misc}. }
    \label{fig:perf-breakdown}
\end{figure}
}
\hide{
\begin{figure}[!t]
    \centering
    \includegraphics[scale=0.27]{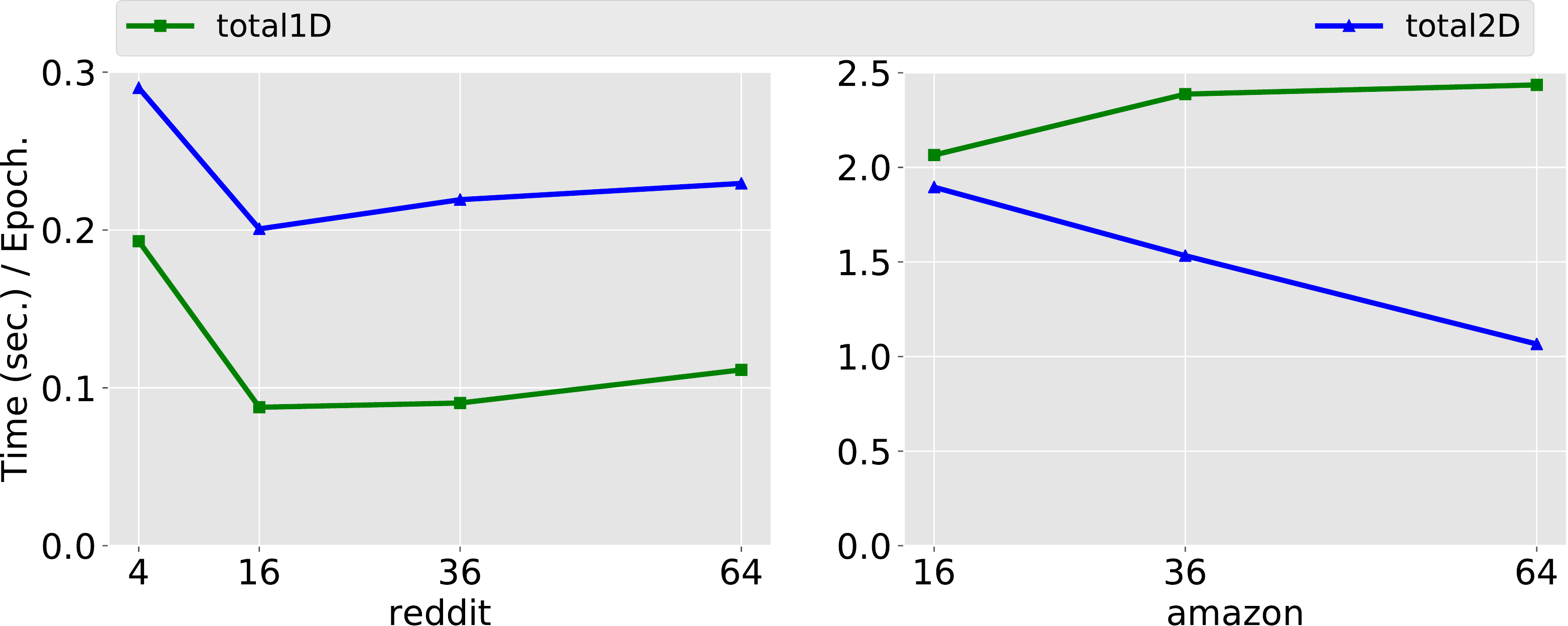}
    \caption{Runtime of 1D Implementation vs. 2D Implementation across GPU counts}
    \label{fig:1dv2d}
\end{figure}
}
\begin{figure}[!t]
    \centering
    \includegraphics[scale=0.27]{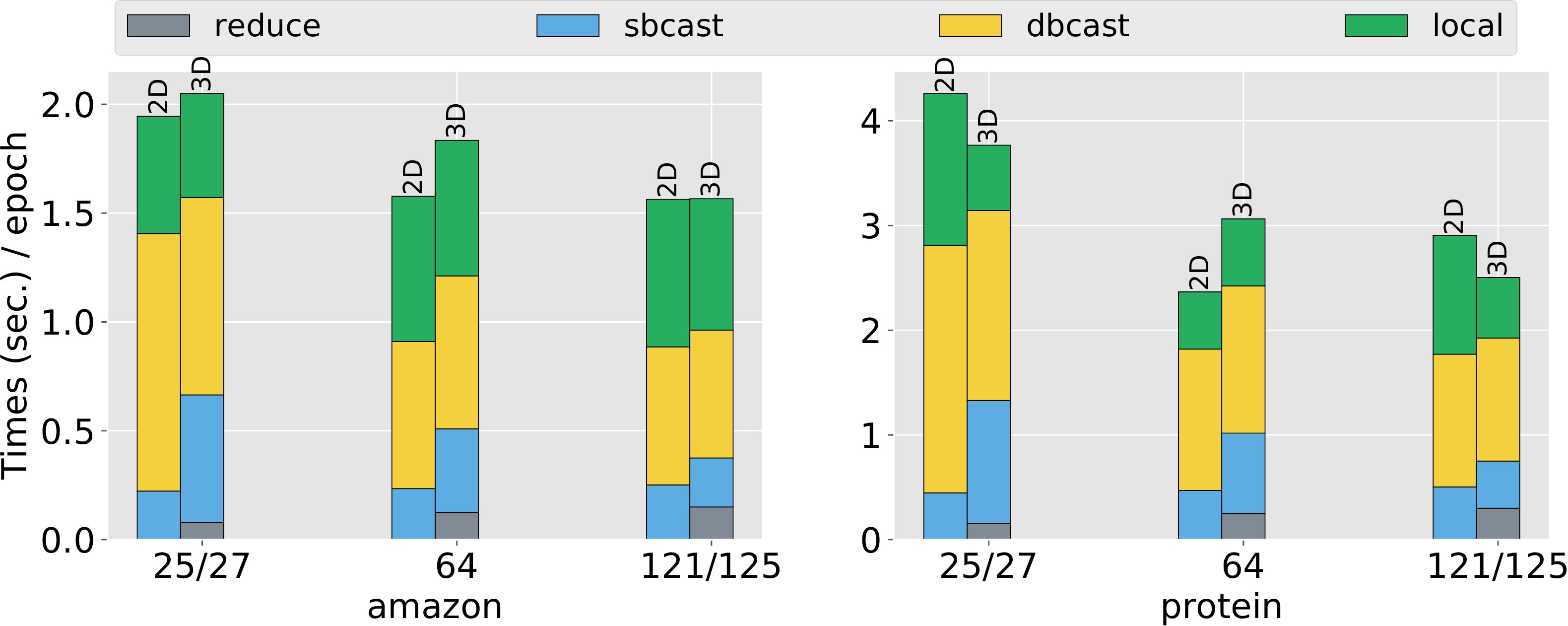}
    \hide{\caption{Runtime of 2D Implementation vs. 3D Implementations across GPU counts. \textit{comm} refers to all communication done by the algorithm, i.e. all broadcasts, reductions. This includes tranposes. \textit{comp} refers to the remainder of the algorithm (i.e. all non-communication).}}
    \caption{Runtime of 2D Implementation vs. 3D Implementations across GPU counts. The x-axis for each subplot is the number of GPUs used. 2D was run on perfect square process counts, while 3D was run on perfect cubes. \textit{reduce} is the allreduce (only for 3D), \textit{dbcast} refers to the broadcast of dense embedding matrices, \textit{sbcast} refers to the broadcast of the sparse adjacency matrix, and \textit{local} is the local computation including cuSPARSE SpMM calls, small DGEMM calls, transpose, and sparse matrix assembly after communication.}
            \vspace{-0.5em}
    \label{fig:2dv3d}
\end{figure}

\subsection{Performance of the 2D Implementation}
\revisiontwo{
The performance and scaling of our 2D implementation is covered by both Figure~\ref{fig:perf-breakdown} (comparing with the 1D and 1.5D implementations) and Figure~\ref{fig:2dv3d} (comparing with the 3D implementation). Figure~\ref{fig:perf-breakdown} covers the process counts $p{=}16,36,64,100$ and uses the default 16 middle layer dimension.  Figure~\ref{fig:2dv3d} covers the process counts $p{=}25,64,121$ and uses 64 as the middle layer dimension. When comparing our 2D and 3D implementations in Figure~\ref{fig:2dv3d}, we used a middle layer dimension of $64$ instead of $16$ because the 3D algorithm needs to partition dimension $16$ across $P^{2/3}$ processes in some cases, which is impossible with $P{>}64$. Consequently, the average feature length $f$ is $\approx 10\%$ larger in Figure~\ref{fig:2dv3d}, translating into a marginal runtime difference between Figures~\ref{fig:perf-breakdown} and~\ref{fig:2dv3d} for $p{=6}4$. \revision{We note that \emph{dbcast} is the total time to broadcast dense matrices in any step of the 2D algorithm and is not restricted to distributed \texttt{GEMM} calls. In fact, \emph{dbcast} time is dominated by the time it takes to communicate  dense matrices (e.g., $\mH$ and $\mG$) during distributed SpMM calls.}}

\revisiontwo{
In terms of scaling with increasing process counts, the 2D algorithm shows noteworthy speedups up until 36 GPUs on all three datasets. Beyond that, Reddit shows slowdowns whereas Amazon and Protein stagnates. For Amazon and Protein, \emph{dbcast} time continues scaling but that is offset by the increase in local computation and/or \emph{sbcast} times.}  There are two fundamental reasons why local SpMM does not scale with increasing process counts. (1) SpMM performance degrades as the matrix gets sparser. Yang et al.~\cite{yang2018design} demonstrate that when the average number of nonzeros per row (i.e., degree, $d = \dnnz / n)$ goes down from 62 to 8, the sustained GFlops rates are cut by a factor of 3. They specifically evaluated cuSPARSE's \texttt{csrmm2} function, which is the same SpMM function we use, but the performance degradation due to increased sparsity is present for all SpMM implementations they evaluated. The average number of nonzeros per row goes down when a sparse matrix is 2D partitioned across larger device counts. This phenomenon is known as hypersparsity~\cite{buluc2008representation} and decreases the average degree of 2D partitioned submatrices by a factor of $\sqrt{p}$. (2) Since the dense matrices (e.g., the activation matrix) are also 2D partitioned, the number of columns in each local dense submatrix also goes down by a factor of $\sqrt{p}$, making the dense matrix ``skinnier''. The performance degradation at this extremely skinny regime is also well documented~\cite{ipdps14}.

\revisiontwo{
In terms of absolute performance, the 2D algorithm is never faster than the 1.5D algorithm. This might come surprising given the asymptotically better communication scaling per analysis in Section~\ref{sec:total2D}. However, even when considering \emph{dbcast} costs alone, the constants in the 2D algorithm are $4\times$ larger than the 1D and 1.5D algorithms. In addition, the 2D algorithm has to communicate the sparse matrix, which has packing and unpacking costs in addition to the costs associated with actually moving the data.}

\hide{
\paragraph{Amazon} For the data we were able to collect on Amazon, we see that the time spent on communicating dense matrices goes down by $2\times$ given $4\times$ more devices. In Figure \ref{fig:perf-breakdown}, this is denoted by ``dcomm''. This is consistent with the bounds discussed in Section \ref{sec:2d} as communication should scale by a factor of $\sqrt{P}$. \revision{We note that \emph{dcomm} is the total time to communicate dense matrices in any step of the algorithm and is not restricted to distributed \texttt{GEMM} calls. In fact, \emph{dcomm} time is dominated by the time it takes to communicate  dense matrices (e.g., $\mH$ and $\mG$) during distributed SpMM calls.}}

%Overall, Amazon sees a $1.8X$ improvement given $4X$ more processes in epoch throughput, as shown in Figure \ref{fig:epoch-through}. The reason the speedup is limited to $1.8X$ is twofold. 

\hide{
We observe that local SpMM does not scale with increasing process count. There are two fundamental reasons for this: 
\begin{enumerate}
    \item SpMM performance degrades as the matrix gets sparser. Yang et al.~\cite{yang2018design} demonstrates that when the average number of nonzeros per row (i.e., degree, $d = \dnnz / n)$ goes down from 62 to 8, the sustained GFlops rates are cut by a factor of 3. They specifically evaluated cuSPARSE's \texttt{csrmm2} function, which is the same SpMM function we use, but the performance degradation due to increased sparsity is present for most of the SpMM implementations they evaluated. Not only is Amazon already sparse with an average degree $24$, but it gets even sparser as its sparse adjacency matrix is 2D partitioned across larger device counts. This phenomenon is known as hypersparsity~\cite{buluc2008representation} and decreases the average degree of 2D partitioned submatrices by a factor of $\sqrt{p}$. 
    \item Since the dense matrices (e.g., the activation matrix) are also 2D partitioned, the number of columns in each local dense submatrix also goes down by a factor of $\sqrt{p}$, making the dense matrix ``skinnier''. For example, the SpMM calls corresponding to middle layer go from multiplying with a dense matrix that has 16 columns (when $p=1$) to multiplying with a dense matrix with only 2 columns (when $p=64$). The performance degradation at this extremely skinny regime is also well documented~\cite{ipdps14}.  
\end{enumerate}
}
\hide{
These two factors have a multiplicative detrimental impact on the local SpMM performance. Future work will involve using more sophisticated SpMM implementations, such as the merge-based method~\cite{yang2018design}, that are less impacted by these factors. Also, wider networks are known to improve testing accuracy of GNNs~\cite{mlsys2020_83} so we expect a trend towards larger number of activations in hidden layers in the future, potentially making the skinny dense matrix issue less relevant. }

\hide{We also note that sparse communication (``scomm'' in Figure \ref{fig:perf-breakdown}) does not scale. This is again due to Amazon's sparsity. Sparse matrix communication here ends up being latency-bound. Furthermore, \revision{as shown by the total communication analysis done in Section~\ref{sec:total2D}, each epoch requires $3\sqrt{P}$ sparse broadcasts. Each of these sparse broadcasts take less than $1$ms at $p=36$ processes}. On the Summit supercomputer, inter-node communication is latency-bound at that point \revision{and further reductions on the data volume can not compensate for the increased latency due to the $\sqrt{P}$ term}. }

\hide{Fortunately, we still see an overall speedup $1.8\times$ when going from 16 to 64 processes in epoch throughput \todo{cite exact numbers}. This is because the most costly operation in training on the Amazon dataset is the communication of dense matrices (such as activation matrices, its derivatives, and intermediate products). Our communication analysis in Section~\ref{sec:2d} accurately predicts this bottleneck where the number of words moved due to communicating dense matrices is more than a factor of 2 larger than the number of words moved due to communicating sparse matrices. The difference is amplified in a multiplicative way by the difference in the average feature vector length ($f\approx 113$) and the average degree ($d\approx24$).}
%The communication of dense matrices in distributed SpMM routines get faster by $2.0\times$ from $16$ to $64$ processes, which is exactly $\sqrt{P}$ factor uncovered by our analysis.

\hide{
\paragraph{Reddit} For the Reddit dataset, we see in Figure \ref{fig:perf-breakdown} that SpMM scales by $5.23\times$ from $4$ to $64$ processes. However, communication does not scale as device count increases. This is because, like the sparse matrix communication in Amazon, communication in Reddit ends up being latency-bound. Additionally, like the situation with Amazon, the average cost of a broadcast is roughly $1$ms. Broadcasts this cheap on Summit end up being latency-bound.}

\hide{
\paragraph{Protein} For our protein network experiments, we see performance improvements on two fronts. First, from $36$ to $100$ processes, the total communication goes down by roughly $1.65\times$. This is consistent with the bounds derived earlier as the communication should scale by a factor of $\sqrt{P} = 10/6$. Second, the SpMM time goes down by roughly $1.33\times$ from $36$ to $100$. This speedup is limited for the same reasons discussed above. Specifically, the performance of SpMM degrades as the matrix gets sparser. In the case of our protein network on $100$ processes, the average degree $\dnnz/n$ becomes roughly $12$, a small fraction of the length of the row.}

\subsection{Performance of the 3D Implementation}
\revision{We run both 2D and 3D implementations on \revisiontwo{$20$} epochs for Amazon and $10$ epochs for Protein. For 2D, we run on process counts $25$, $64$, and $121$. For 3D, we run on process counts $27$, $64$, and $125$. We specifically choose these counts to permit a fair comparison. We also do not run the 3D algorithm on Reddit as it was already latency-bound and, thus, cannot leverage the benefits of the 3D algorithm.}

% \revisiontwo{\paragraph{Amazon:} We see in Figure \ref{fig:2dv3d} that for lower process counts, 2D tends to outperform the 3D implementation overall by roughly $15$\%. Between $P=25/27$ and $P=121/125$, we see that the}
\revisiontwo{Overall, we see that the 3D implementation enjoys a consistent speedup on both datasets all the way until 125 GPUs, unlike the 2D algorithm.
While the 3D algorithms is slower than the 2D for smaller process counts on Amazon, it matches the 2D implementation when $P=121/125$. For Protein, the 3D algorithm generally outperforms 2D. The difference here lies in the computation. With 2D, as discussed above, scaling computation to large process counts is difficult since the $\dnnz$ per row shrinks as $P$ increases. While in 2D, each row is partitioned across $P^{1/2}$ processes, in 3D they are partitioned across $P^{2/3}$ processes. Thus, the average $\dnnz$ per row is lower in 3D than in 2D.}

\revisiontwo{For both datasets, 3D communication scales roughly $5-10\%$ better than 2D, to the point where they both perform equally well at $P=121/125$.} While this is a small difference, this trend is consistent with the analysis in Sections \ref{sec:2d} and \ref{sec:3d}. While the bandwidth in 2D scales by a factor of $P^{1/2}$, the bandwidth in 3D scales by $P^{2/3}$. The 3D analysis also has larger constant factors. Coupled together, we expect the 3D implementation to scale only slightly better than the 2D implementation. \revisiontwo{We expect that, for larger process counts, the 3D implementation will leverage additional scaling to outperform 2D for Amazon}.

\hide{\revision{We see that the 3D implementation runs slower than 2D, between $17\%-30\%$ slower on large process counts in Figure \ref{fig:2dv3d}. This difference is due to computation not scaling and occasionally slowing down (as is the case with Amazon). The difference is likely because the difficulties scaling computation in the 2D implementation are magnified in 3D. With 2D, as discussed above, scaling computation to large process counts is difficult since the $\dnnz$ per row shrinks as $P$ increases. While in 2D, each row is partitioned across $P^{1/2}$ processes, in 3D they are partitioned across $P^{2/3}$ processes. Thus, the average $\dnnz$ per row is lower in 3D than in 2D. Amazon, already sparse, suffers from this phenomenon more than Protein.}}

\hide{\revision{Between 2D and 3D, however, we do see that the 3D communication scales slightly better than 2D. Across all process counts in Figure \ref{fig:2dv3d}, 3D has between $5-10\%$ better improvement than 2D. While this is a small difference, this is consistent with the analysis in Sections \ref{sec:2d} and \ref{sec:3d}. While the bandwidth in 2D scales by a factor of $P^{1/2}$, the bandwidth in 3D scales by $P^{2/3}$. The 3D analysis also has larger constant factors. Coupled together, we expect the 3D implementation to scale only slightly better than the 2D implementation. We hypothesize that for the 3D algorithm to beat 2D, we would need large process counts to overcome the issues with computation scalability.}}

\hide{
\subsection{1D vs. 2D Implementations Analysis}
\revision{When comparing our 1D and 2D implementations, we run both the Reddit and Amazon with a middle layer dimension of $16$. We also run both implementations on $100$ epochs.}

\revision{The main takeaway from this comparison is that for smaller datasets, a 1D algorithm is likely the better choice. Note that 1D Reddit runs roughly $2\times$ faster than 2D Reddit as seen in Figure \ref{fig:1dv2d}. For Amazon, 1D and 2D are comparable on low process counts, but 2D Amazon is $2.5\times$ faster than 1D on $64$ processes. We know that 2D for Reddit does not scale well as it quickly becomes latency-bound. Thus, Reddit is only worth running on smaller process counts. For these counts, though, 1D algorithms typically outperform 2D due to low constant factors. This explains why 1D for Reddit outperforms 2D. We do not see Amazon 1D outperforming 2D on low process counts simply because running Amazon on low process counts runs out of memory. Recall that the communication of the 1D implementation does not scale with increasing process counts, while 2D scales by a factor of $P^{1/2}$. Since Amazon 1D and 2D are comparable on $16$ processes, we hypothesize that Amazon 1D would outperform Amazon 2D on small process counts given more memory per GPU.}
}

\hide{
\subsection{Comparisons}  However, we can do an indirect comparison, even though our exact hardware configuration is different than theirs. Our 1.5D GNN training achieves $xx$ epochs/sec throughput on Reddit on 16 GPUs, which is more than 2X faster than ROC. 
\Aydin{Say that our Reddit epochs per second is faster than anything published, including ROC and Neugraph, even though the hardware configurations are different.}}

\hide{Comparing reported numbers is not meaningful because we do not have access to a hardware system that is equipped with the same configuration as theirs. More importantly, our analysis predicts that our 2D implementation will only be competitive with 1D approaches when $\sqrt{p} \geq 5$. Given that the largest GPU counts reported by Neugraph and ROC are 8 and 16, respectively, the benefits of our 2D implementation would not be apparent. We would like to highlight that our experiments are run on significantly larger GPU counts than these contemporary studies. }

\section{Conclusions and Future Work}
We presented distributed-memory parallel GNN training algorithms that \revisiontwo{asymptotically reduce communication costs by dividing two or three dimensions (1.5D, 2D, or 3D) of the iteration space across the training pipeline when compared to the commonly used (1D) vertex partitioning methods.  We evaluated these algorithms on three datasets that differ in graph size and structure, revealing a set of trade-offs in memory use, local node efficiency, and communication volume, latency (number of messages), and efficiency.}

\revisiontwo{
Our experimental results show that even simple variants of our algorithms can be scalable when implemented using off-the-shelf tools such as PyTorch and cuSPARSE. On a flat execution model that uses only a single GPU per node, there are clear benefits to the communication avoiding approach. On the other hand, the memory hierarchy on a six-GPU node leads to some surprising results due to issues such as injection bandwidth limitations, and the anomalies suggest opportunities to improve collective communication through hierarchical algorithms so that compute nodes' injection bandwidth to the network does not become a bottleneck. Overall, we show that the 1.5D algorithm, which can scale with available memory in the second dimension, is the most effective given the implementation environment on the Summit machine.  We believe our communication-avoiding algorithms will show their true benefit on better connected architectures such as Perlmutter~\cite{perlmuttertalk} where each GPU gets its own $25\,$ GB/s injection bandwidth. At that point, new local SpMM implementations whose throughput degrade more gracefully with increasing sparsity will be crucial for scaling the computation part of GNN training.}

The memory consumption of GNNs can be high due to the need to store activations generated during forward propagation so they can be used for backpropagation. If full-batch gradient descent is used or if graph sampling does not provide an asymptotic reduction, the memory costs become $O(nfL)$, which is prohibitive for deep networks. Consequently, we envision future work where our distributed training algorithms are carefully combined with sophisticated sampling based methods to achieve the best of worlds.

\section*{Acknowledgments}
This material is based upon work supported by the National Science Foundation
Graduate Research Fellowship under Grant No. DGE 1752814 and by the National Science Foundation 
under Award No. 1823034. This work is also supported in part by the Advanced Scientific Computing Research (ASCR) Program of the Department of Energy Office of Science under contract
No. DE-AC02-05CH11231, and in part by the Exascale Computing Project
(17-SC-20-SC), a collaborative effort of the U.S. Department
of Energy Office of Science and the National Nuclear Security
Administration.

This research used resources of the Oak Ridge Leadership Computing Facility at
the Oak Ridge National Laboratory, which is supported by the Office of Science
of the U.S. Department of Energy under Contract No. DE-AC05-00OR22725.

\bibliographystyle{plain}
\bibliography{cagnn}
\end{document}